\documentclass[runningheads]{llncs}

\usepackage{eccv}

\usepackage{eccvabbrv}
\usepackage{wrapfig}
\usepackage{float} 
\usepackage{caption}

\usepackage{graphicx}
\usepackage{booktabs}
\usepackage{fancyvrb}
\usepackage[accsupp]{axessibility}  

\usepackage{hyperref}

\usepackage{orcidlink}

\begin{document}


\title{Finding Visual Task Vectors}


\author{\small Alberto Hojel\inst{1}\and
Yutong Bai\inst{1}\and
Trevor Darrell\inst{1}\and
Amir Globerson\inst{2,3} \and
Amir Bar\inst{1,2}}


\authorrunning{A. Hojel et al.}

\institute{UC Berkeley \and
Tel Aviv University \and Google Research}
\institute{\quad\quad
    $^1$UC Berkeley \quad 
    $^2$Tel Aviv University \quad 
    $^3$Google Research
}

\maketitle
\vspace{-5mm}
\begin{figure}
\centering
\includegraphics[width=1\textwidth]{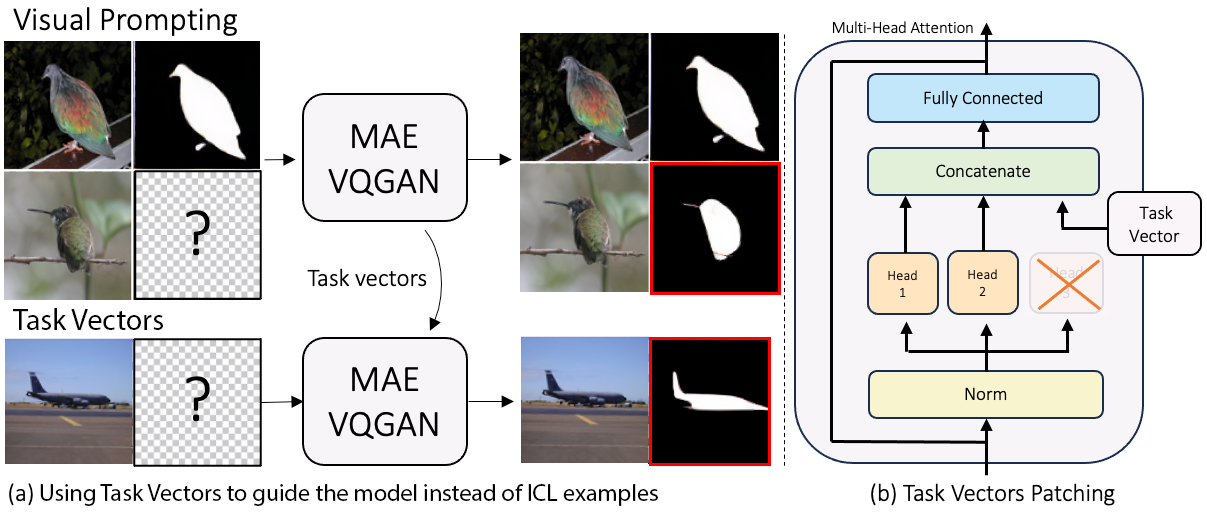}
\captionof{figure}{Visual Prompting models like MAE-VQGAN~\cite{bar2022visual} require input-output example(s) to describe the desired task in their forward pass. We analyze the model activations and find \emph{Task Vectors}, activations that encode task information that can be reused to control the task the model performs (see Figure~\ref{fig:teaser}a). Specifically, we tap into activations of individual attention heads and replace their outputs with Task Vectors to guide the model to the desired task (see Figure~\ref{fig:teaser}b). Surprisingly, the resulting models perform better than the original model while removing the need for input-output examples. This confirms that \emph{Task Vectors} exist in the network activation space and they can guide the model to perform the desired task.}
\label{fig:teaser}
\vspace{-5mm}
\end{figure}

\begin{abstract}
\vspace{-5mm}
Visual Prompting is a technique for teaching models to perform a visual task via in-context examples, without any additional training. In this work, we analyze the activations of MAE-VQGAN, a recent Visual Prompting model~\cite{bar2022visual}, and find \textit{Task Vectors}, activations that encode task-specific information. Equipped with this insight, we demonstrate that it is possible to identify the Task Vectors and use them to guide the network towards performing different tasks without having to provide any in-context input-output examples. To find Task Vectors, we compute the mean activations of the attention heads in the model per task and use the REINFORCE~\cite{williams1992simple} algorithm to patch into a subset of them with a new query image. The resulting Task Vectors guide the model towards performing the task better than the original model.\footnote{For code and models see~\url{www.github.com/alhojel/visual_task_vectors}}
\end{abstract}

\section{Introduction}
\label{sec:intro}
In-context learning (ICL) is an emergent capability of large neural networks, first discovered in GPT-3~\cite{brown2020language}, which allows models to adapt to novel downstream tasks specified in the user's prompt. In computer vision, Visual ICL (known as Visual Prompting~\cite{bar2022visual,bahng2022exploring,bai2023sequential,zhang2024makes}) is still at its infancy but it is increasingly becoming more popular due to the appeal of using a single model to perform various downstream tasks without specific finetuning or change in the model weights.

In this work, we ask how in-context learning works in computer vision. While this question has yet to be explored, there has been a significant body of research in Natural Language Processing (NLP) trying to explain this phenomenon~\cite{akyurek2022learning,dai2022can,garg2022can,hahn2023theory,han2023context}. Most recently, Hendel et al.\cite{hendel2023context} suggested that LLMs encode \textit{Task Vectors}, these are vectors that can be patched into the network activation space and replace the ICL examples while resulting in a similar functionality. Concurrently, Todd et al.~\cite{todd2023function} discovered \textit{Function Vectors}, activations of transformer attention heads that carry task representations. Our work is inspired by these observations and we aim to study ICL in computer vision.

Do Task Vectors exist in computer vision models as well? To build intuition, we start by exploring the activation space of MAE-VQGAN~\cite{bar2022visual}. Intuitively, we look for intermediate activations that are invariant to changes within a task, but have high variance across different tasks. We use this simple and fast-to-compute metric to rank activations according to their ``taskness''. 
Using this score, we find that the activation space of certain attention heads can perfectly cluster the data by tasks, hinting at the existence of \textit{Visual Task Vectors}.

Based on this insight, we hypothesize that Visual ICL models create Task Vectors too, and aim to find them. However, finding Visual Task Vectors by relying on previous approaches is challenging. For example, both~\cite{hendel2023context,todd2023function} restricted their search space to the output activations of the last token in the prompt sequence (colored in red in the following example: 
\begin{BVerbatim}[commandchars=\\\{\}]
``Banana:B, Apple\textcolor{red}{:}''
\end{BVerbatim}
). This approach is natural for text as it is processed sequentially by autoregressive models like LLaMA~\cite{touvron2023llama}, but with images (see Figure~\ref{fig:teaser}a, ``Visual Prompting'' input image), it is not obvious what token activations hold task information, and architectures like MAE-VQGAN~\cite{bar2022visual} do not process image tokens sequentially. This alone increases the search space significantly because multiple tokens might hold Task Vectors. 
Additionally, while Hendel et al.~\cite{hendel2023context} and Todd et al.~\cite{todd2023function} differ in the activation space in which they perform interventions, both methods evaluate single interventions, ignoring emergent second-order effects.

To identify Task Vectors, we first compute the mean attention head outputs for each token position across a set of task examples (Figure~\ref{fig:teaser}a). We then replace attention head outputs at selected positions with these pre-calculated means, using a new query as input. Each position is modeled as a random variable, and we use REINFORCE~\cite{williams1992simple} to optimize patching positions, guiding the model towards the desired task. This approach yields competitive performance across various tasks, validating the existence of Task Vectors. Our method effectively explores higher-order effects and complex interactions between attention heads in different layers, improving upon previous approaches that examine activations in isolation.

Our contributions are as follows. We show evidence for the existence of \textit{Visual Task Vectors} and propose a practical way to identify them. Moreover, we find that patching the resulting Task Vectors guides the model towards the desired task with better performance compared to the original in-context prompts while reducing the number of FLOPS needed for a forward pass by $22.5\%$.

\section{Related Work}
\label{sec:rw}

\subsection{Visual Prompting}
Visual Prompting~\cite{bar2022visual,bahng2022exploring,jia2022visual,xu2023improv,zhang2024makes,bai2023sequential} is a class of approaches to adapt computer vision models to downstream tasks, inspired by the success of prompting in NLP~\cite{brown2020language}. Approaches like~\cite{bahng2022exploring,jia2022visual} seek to improve task-specific performance by adding trainable prompt vectors to the model. Other Visual Prompting approaches allow a model to handle various vision tasks~\cite{bar2022visual,xu2023improv,bai2023sequential,zhang2024makes} by introducing visual examples or text at the time of inference. Such prompting is related to the way in-context learning~\cite{xie2021explanation, wei2022chain,liu2021makes, lu2021fantastically} operates in language models~\cite{radford2019language, brown2020language,wang2021gpt}. In fact, trainable prompts and in-context learning can be viewed as two complementary approaches for ``describing'' a task to a model~\cite{li2021prefix}.
Our goal here is to better understand the underlying mechanism of Visual ICL, and we analyze the MAE-VQGAN model presented in~\cite{bar2022visual}.

\subsection{Explainability}
Causal Interventions \cite{bau2018gan, park2023linear, pearl2022direct, meng2022locating} and Activation Patching~\cite{zhang2023towards} are valuable tools for understanding complex neural networks' internal mechanisms, enhancing model interpretability \cite{zhang2021survey, moraffah2020causal, singh2024rethinking}. These methods enable systematic examination of how models encode information and represent high-level concepts \cite{zhang2024batch, wu2024transformer, lu2023prompts}. By manipulating internal states or inputs and observing output changes, they reveal causal structures and effects driving model predictions \cite{gandelsman2023interpreting}. In this work, we employ Activation Patching~\cite{zhang2023towards} to improve Visual Prompting models' guidance for various computer vision tasks through targeted interventions.

\subsection{Task Vectors}

In~\cite{hendel2023context,todd2023function,ferry2023emergence}, Task Vectors or Function Vectors are sets of latent activations derived from particular positions inside a transformer~\cite{attention} model which serve as internal representations of a task implicitly described by an ICL prompt with input-output demonstrations. These latent activations at particular positions can consequently be used in a new forward pass in the absence of the ICL prompt (or with a corrupted prompt) while still managing to guide the model to perform the desired task.
The investigation of Task Vectors aligns with broader efforts in the field to make neural networks more adaptable and tailored to specific tasks~\cite{liu2023context, luo2024understanding} as well as boosting the performance~\cite{palit2023towards, xu2024exploring, jin2024cutting} by gaining a deeper understanding of how different attention heads within a model contribute to its overall function. Our work is the first to explore Task Vectors in computer vision.

\section{Methods} \label{sec:methods}
Our goal is to understand in-context learning for computer vision and how existing models can be adapted to different downstream tasks at inference time. We focus on the MAE-VQGAN model \cite{bar2022visual}, a variant of MAE \cite{DBLP:journals/corr/abs-2111-06377} with a Vision Transformer \cite{dosovitskiy2021image} encoder-decoder architecture.
Given an input-output example $(x_s, y_s)$ and a new query $x_q$, to succeed in an ICL task, the model $F$ must implicitly apply the transformation from $x_s$ to $y_s$ over $x_q$ to produce $y_q$:
\begin{equation} \label{eq:prompting}
y_q = F(x_s, y_s, x_q)
\end{equation}

Based on observations from NLP \cite{hendel2023context,todd2023function}, we hypothesize that computer vision models also encode latent Task Vectors in their activation space during the forward pass. This requires the model to implicitly map the ICL example into a set of latent Task Vectors $z = \{z_i\}^{k}_{i=1}$ which we derive from the attention head outputs for different tokens across the model's layers (the internal representations). The original function $F$ can then be decomposed into extracting the Task Vectors by a function $G$ and applying $F$ on the query while fixing the computed task activations $z \in \mathbb{R}^d$, where $d$ is the hidden dimension of the model.
\begin{align}
z &= G(x_s, y_s) &
y_q &= F(x_q|z)
\end{align}

We now describe how Task Vectors can be derived from the internal representations of a model.

\subsection{Computing Mean Activations} \label{sec:computing_mean_acc}
Let $i=(l,m,k)$ denote the position in the model, where $l$, $m$, and $k$ are the attention block, head, and token indices respectively. Define $H_i: (x_s, y_s, x_q) \rightarrow \mathbb{R}^d$ as the function that outputs the activation at position $i$ for a given demonstration and query triplet. Let $(x_{s}, y_{s}, x_{q}) \sim \mathbf{D_{task_j}}$ be a triplet of input-output example and a query from task $j$. We compute the intermediate activation $h^{i}_{task_j}$ via $H_i$ and denote its mean activation as $\mu_{i,j}$:
\begin{align}
h^{i}_{task_j} = H_{i}(x_{s}, y_{s}, x_{q}) && \mu_{i,j} = \mathbb{E}[h^{i}_{task_j}]
\end{align}


\subsection{Scoring Activations} \label{sec:activation_scoring}
Intuitively, every task vector is an activation that changes across different tasks but remains relatively invariant to changes within a specific task. We define the data distribution $(x'_s, y'_s, x'_q) \sim \mathbf{D_{all\_tasks}}$ as the union of all task-specific distributions, and the intermediate activations $h^{i}_{all} = H_{i}(x'_{s}, y'_{s}, x'_{q})$. The scoring function $\rho_{token}(i)$ is defined as the ratio of inter-task to intra-task variance as $\rho_{token}(i)$ where $\text{Var}(\cdot)$ denotes the variance, $h[e]$ is the $e$-th element of vector $h$, and $n$ is the number of tasks.
\begin{align}
\label{eq:rho_mu}
\rho_{token}(i) &= \frac{\sum_{e=1}^d \text{Var}(h^i_{all}[e])}{\frac{1}{n}\sum_{j=1}^n \sum_{e=1}^d \text{Var}(h^i_{task_j}[e])}
\end{align}

\begin{figure}[p]
    \centering
    \centering
    \includegraphics[height=0.98\textheight]{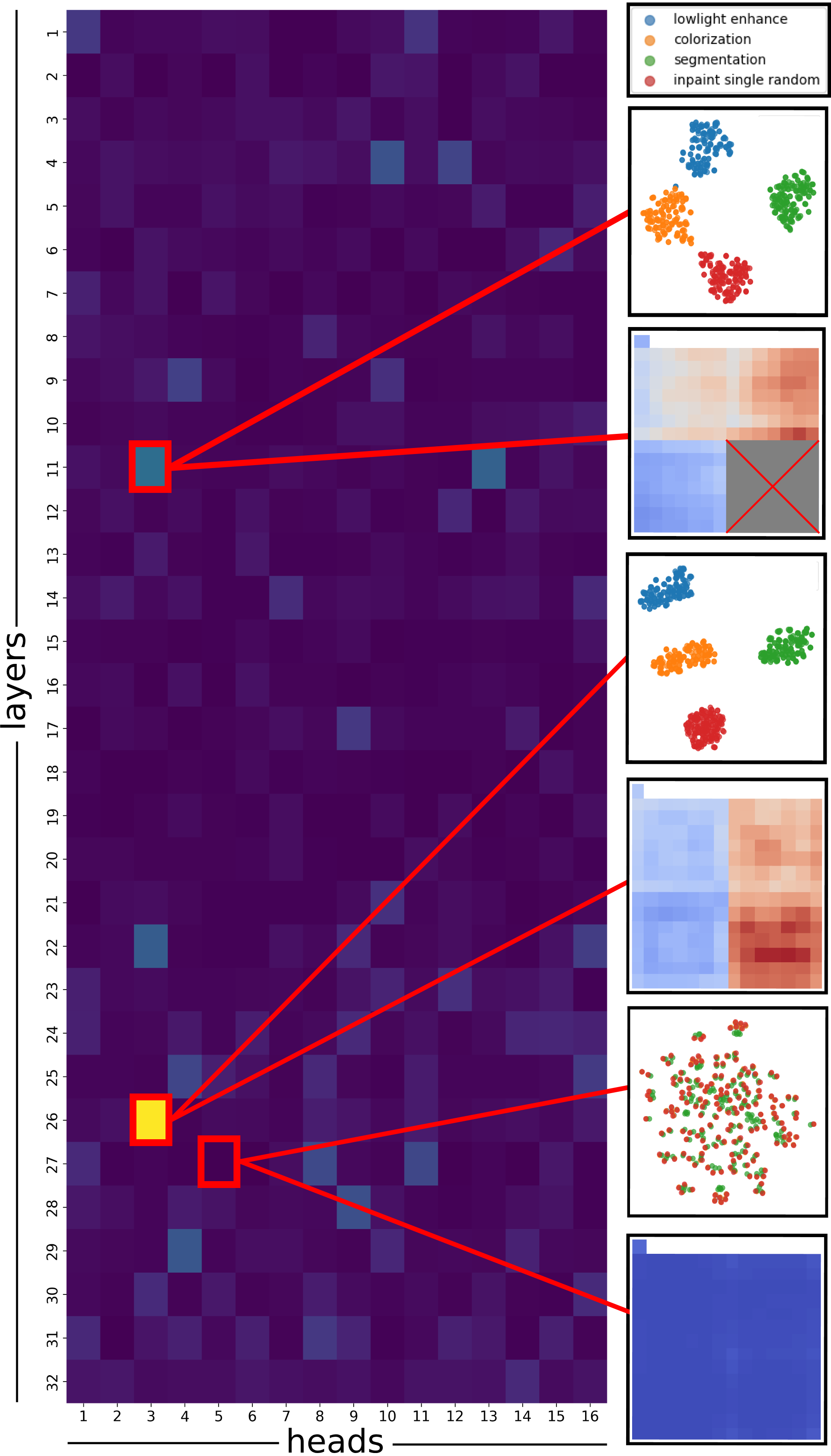}
    \caption{\textbf{Activation Scoring Analysis.} Individual scores (${\rho_{token}}(i)$) aggregated per Attention Head (left) for the encoder and decoder. Individual token scores of specific heads, along with the t-SNE~\cite{JMLR:v9:vandermaaten08a} clustering of head activations of different tasks (right).}
    \label{fig:headscores}
\end{figure}

Computing these is efficient, requiring only forward passes across batches of data. This scoring function identifies "taskness" in activations (see Figure \ref{fig:headscores}, and Table \ref{tab:metrics_comparison}), but may also highlight other task-variant activations (e.g., color histograms varying across tasks but consistent within tasks like binary segmentation maps versus colored images).
We use these scores mainly to build intuition, analyze task clusters in the activation space (Section \ref{sec:results}), and develop a simple baseline (``Greedy Random Search'', see Suppl.~\ref{sec:grs_impl}). However, we employ a more robust general approach to identifying high-order Visual Task Vectors, as described in the following section.

\subsection{Finding Visual Task Vectors via REINFORCE} 
\label{sec:reinforce_methodology}
How can Task Vectors be identified? An exhaustive search over all subsets of activations is intractable. For example, MAE-VQGAN~\cite{bar2022visual} which we utilize here has $32$ attention blocks, each with $16$ heads, therefore, if we consider these to be the potential set of activations then we need to search over $2^{512}$ options, evaluating each option over a held-out validation set.

For every task $j$ we can apply the following procedure to find the Task Vectors. Recall that $\{\mu_{i,j}\}$ denotes the mean activations and $F(\cdot)$ is a pretrained visual prompting model. Denote $\alpha_{i,j} \sim Bernoulli(\sigma(\theta_{ij}))$ as random variables that signify whether the mean task activation $\mu_{i,j}$ is a task vector of task $j$ placed in activation position $i$ and $\theta_{ij}$ is a learned weight followed by the sigmoid function to ensure it is in $[0,1]$.

Denote $z_j$ as the set of Task Vectors for task $j$: $z_j = \{\mu_{i,j} | \alpha_{i,j}=1\}$. Given pairs of input-output task demonstrations $x_q, y_q \sim \mathbf{D_{task_j}}$,\footnote{We overload the definition of $\mathbf{D_{task_j}}$ to avoid notation clutter.} we want to find the set of Task Vectors that minimizes the loss function:
\begin{equation}
L(\theta) = E_{{z_j} \sim p_\theta} L(y_q, F(x_q|z_j))
\end{equation}
Where $p_\theta$ denotes the sampling distribution of $z_j$ and $L$ denotes the loss function that suits the particular task $j$. Note that differently than in Equation~\ref{eq:prompting}, if we find a good set of Task Vectors $z_j$, then we no longer need to condition $F$ over additional input-output examples.

To find a set of Task Vectors, we need to estimate the parameters $\theta$. Since the sampling distribution $p_{\theta}$ depends on $\theta$, it is natural to use the REINFORCE algorithm~\cite{williams1992simple}. The key idea in REINFORCE is the observation that:
$$\nabla L(\theta) = E_{z_j \sim p_\theta}{L(y_q, F(x_q|z_j))\nabla \log p_{\theta}(z_j)}$$

Thus, we can approximate $\nabla L(\theta)$ by sampling ${z_j}$ and averaging the above equation. After iteratively optimizing with gradient descent, we select the final task vector positions by sampling a set of $z_j$.

Instead of learning Task Vectors placements ${\alpha_{ij}}$ for each individual task, we can revise the procedure above to learn a single placement by defining random variables $\alpha_{i} \sim Bernoulli(\sigma(\theta_{i}))$ that accommodate all tasks. In this setting, training requires drawing examples across tasks with their respective task-wise mean activations and task specific loss function. We refer to this setting as \textit{multi-task} patching. Furthermore, the search can be applied at different granularities, such as patching token groups from the same quadrant, all tokens in an attention head, or entire layers. We discuss these design choices in the next section.

\section{Experiments}
\label{sec:exps}

Our experiments explore if activation patching of Task Vectors can make a model perform the desired visual task as well as or better than its original one-shot performance. We describe the implementation of MAE-VQGAN, prompting schemes, baselines for comparison, visual tasks, and conducted experiments.

\subsection{Implementation Details}
\noindent\textbf{MAE-VQGAN~\cite{bar2022visual}.} An MAE ~\cite{DBLP:journals/corr/abs-2111-06377} with a ViT-L~\cite{dosovitskiy2021image} backbone.
The decoder predicts a distribution over a VQGAN~\cite{Esser_2021_CVPR} codebook to output images with better visual quality. We used the pretrained checkpoint from~\cite{bar2022visual} where it was trained over the Computer Vision Figures~\cite{bar2022visual} dataset and ImageNet~\cite{russakovsky2015imagenet}.

\noindent\textbf{Finding Task Vectors.} Similar to one-shot (Section~\ref{sec:baselines}), we use a 2x2 image grid with the prompt. For task vectors, we embed only the query in the bottom left quadrant. The model reconstructs only the output part (bottom right). We patchify the query image at 112x112 resolution, apply bottom left quadrant positional encodings, feed patches into the encoder, and process them with the decoder alongside bottom right quadrant mask tokens to obtain the result.

We intervene on attention head outputs by replacing them with mean activations at positions identified using REINFORCE~\cite{williams1992simple}. Initially, we set $\theta_{ij} = -1$. To reduce the search space, we group patching positions into three categories: CLS (1 token), bottom left quadrant (49 query image patch tokens), and bottom right quadrant (49 MASK tokens). In each iteration, we sample 32 times from the Bernoulli distribution for each of 10 images, patching positions where the sampled value is 1. This results in 320 executions of task vector conditioned MAE-VQGAN per iteration. We optimize Bernoulli parameters as outlined in Section~\ref{sec:reinforce_methodology} using Adam~\cite{kingma2017adam} with a learning rate of 0.1. The algorithm runs for 600 steps, selecting the best checkpoint every 50 steps based on evaluation on a held-out test set.

\subsection{Activation Scoring Analysis}
 Here our goal is to evaluate the Activation Scoring step (outlined in Section~\ref{sec:activation_scoring}), specifically whether high scoring activations indeed correspond to Task Vectors.

\noindent\textbf{Collecting Activations.}
To compute activation scores, we first run the model's forward pass in a one-shot setting across different tasks (Section~\ref{sec:computing_mean_acc}). We use 100 prompts and queries from Pascal 5i~\cite{shaban2017one} training set, ensuring reasonable one-shot performance. We save the activations for each task $j$ and position $i=(l,m,k)$, where $l$, $m$, and $k$ are the attention block, head, and token indices respectively. We then compute the mean activation ${\mu_{i,j}}$ and score $\rho_{token}(i)$.

\noindent\textbf{Evaluation via Clustering.}
 Next, we wish to analyze if $\rho_{token}(i)$ indeed captures ``taskness''. Intuitively, we expect layers that capture task information to succeed in clustering activations by task. To assess this, we analyze the clustering performance of vectors with high-ranking activations versus those marked with low scores. We measure the clustering performance using common clustering metrics like the Silhouette Score~\cite{ROUSSEEUW198753}
and the Davies-Bouldin Score~\cite{davies-boulding-79}.
 Finally, we also perform a qualitative analysis by visualizing the representations on a t-SNE~\cite{JMLR:v9:vandermaaten08a} plot, coloring each data point by it's task label. 

\subsection{Downstream Tasks}
We evaluate the performance on standard image-to-image tasks like Foreground Segmentation, Low Light Enhancement, In-painting, and Colorization.

\noindent\textbf{Dataset.} We utilize Pascal-5i~\cite{shaban2017one}, consisting of 4 image splits (346-725 images each) with segmentation masks. For evaluation, We sample 1000 prompt-query pairs per split from the validation set. For Activation Scoring and methods to find Task Vectors, we use only training set examples.
 
\noindent\textbf{Foreground Segmentation.} We use Pascal-5i~\cite{shaban2017one} segmentation masks and report mean IOU (mIOU) across four splits.

\noindent\textbf{Low Light Enhancement.} We multiply Pascal-5i image color channels by 0.5 for input, using the original as output. We report Mean Squared Error (MSE).

\noindent\textbf{Inpainting.} We mask a random 25\% square region (1/8 area) of each image for input. We report MSE using the original image as output. 

\noindent\textbf{Colorization.} To obtain input-output pairs, we convert an image to grayscale and denote it as the input, and have the output be the original image. For evaluation, we report the MSE metric.

\subsection{Baselines}
\label{sec:baselines}
\noindent\textbf{One-shot Prompting.} We follow the basic one-shot setup in~\cite{bar2022visual}. Specifically, we construct a grid-like image structure with an input-output demonstration, a query, and a masked output region which are embedded into a $2x2$ grid. We feed this grid image to the model to obtain the output prediction which we use for evaluation purposes. 

\noindent\textbf{Causal Mediation Analysis.} We compare our methodology with the Causal Mediation Analysis methodology as presented in~\cite{todd2023function} as a baseline. We select the top 25\% of activations with the highest causal score across 10 images.

\noindent\textbf{Greedy Random Search.} We compare our methodology to an iterative greedy random search algorithm (GRS) used to select Task Vectors based on the activation scoring metric proposed in Section~\ref{sec:activation_scoring}. This serves as a baseline and is outlined in the Supplementary Materials (Section \ref{sec:grs_impl}).

 \subsection{Ablations}


In this section, we describe the set of experiments conducted to validate our implementation choices of our REINFORCE~\cite{williams1992simple} method.

\noindent\textbf{Task Vectors Location in Encoder vs. Decoder.}
We hypothesize that task implementation spans both encoder and decoder. To test this, we apply interventions to the encoder only, decoder only, and the whole network. We report mIoU on four Segmentation task splits to assess the necessity of interventions in both parts for effective task implementation.

\noindent\textbf{Patching Granularity.} We explore intervention granularities by grouping token positions to reduce dimensionality and optimization search space. Positions \(i=(l,m,k)\) are grouped into spatial quadrants or attention heads. To balance precision and search space, we use three granularity levels: individual tokens, quadrants, and attention heads, and report performance across four tasks.


    \subsection{Task Vector Patching}
    \label{sec:vector_patching_exp}

We investigate whether patching Task Vectors into a model's forward pass, without ICL demonstrations, can achieve performance comparable to one-shot prompting. Our experiments include:

\noindent\textbf{Task-specific.} For each of the four tasks we execute our task-specific method, and the baselines of Causal Mediation Analysis (CMA) and Greedy Random Search (GRS) with 10 images and report the results on corresponding tasks.

\noindent\textbf{Multi-task.} We apply our method in a multi-task scenario, selecting two images per task and evaluating alongside CMA and GRS baselines. We jointly assess loss across all tasks, normalizing per task to ensure equal weighting and prevent single-task dominance. We include an identity-copy task to maintain a consistent batch size of 10 for fair comparisons.

 \noindent We also provide the following baselines to benchmark the performance of the three algorithms, and to validate the necessity of top $k$ score ranking the layers and performing the iterative search for the GRS method:

\noindent\textbf{MAE-VQGAN.} We compare to the original model's one-shot performance. 

\noindent\textbf{Random Quadrants.} We patch into randomly sampled positions across the whole network (same total amount of patches as task-specific and multi-task)

\noindent\textbf{GRS across Random K Layers.} We execute the search algorithm on a random set of $k$ layers, instead of score-ranked layers, to validate the necessity of our proposed scoring method.

\noindent\textbf{Patching into Top Quadrants.} We patch into the top quadrants based on their scoring (same total amount of patches as task-specific and multi-task). That is, we directly patch into areas with high scores naively without the iterative refining steps of the GRS, to validate its utility.

\begin{figure}[p]
    \centering
    \includegraphics[height=\textheight]{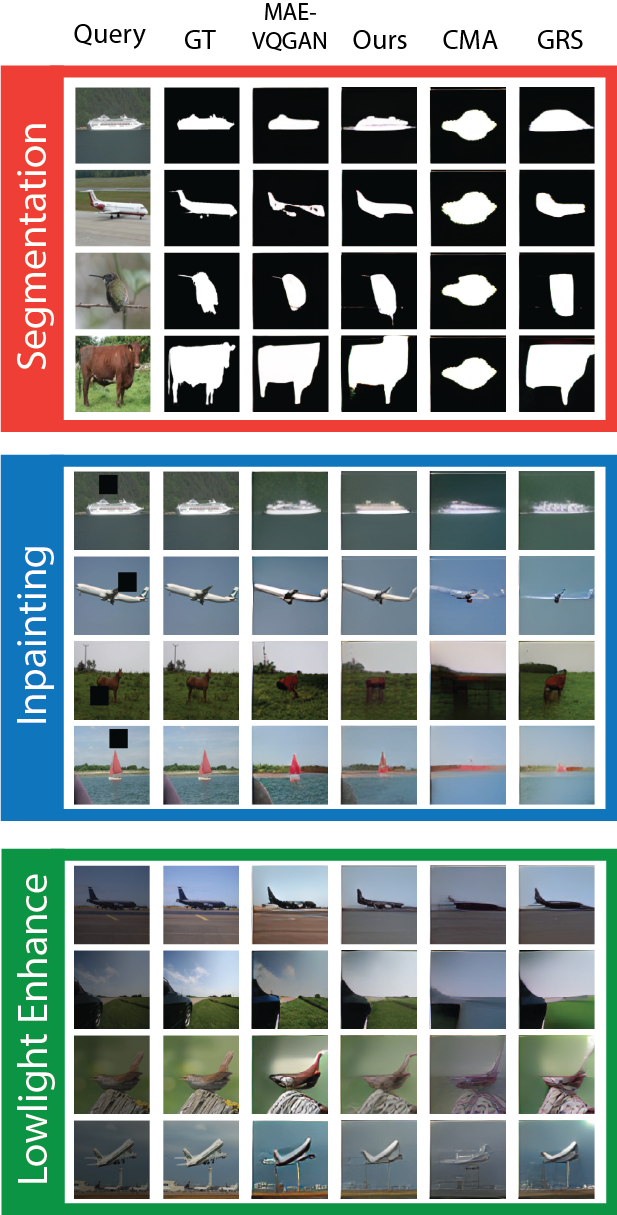}
    \caption{\textbf{Qualitative Examples.} We qualitatively compare the task-specific variants of our methodology's results with the original model and the CMA and GRS baselines. Our patching methodology performs better than the original MAE-VQGAN model.}
    \label{fig:example}
\end{figure}

 We report results across four splits for all four tasks, comparing our variants to the original MAE-VQGAN model and random baselines. Supplementary Materials (Section \ref{sec:exps}) include initial explorations of REINFORCE-powered Language Task Vectors with Llama7B, vector addition to create composable tasks, and others.

\section{Results}
\label{sec:results}

\subsection{Activation Scoring Analysis}

We compute and display ${\rho_{token}}(i)$ per head on a heatmap, with layers on the y-axis and head indices on the x-axis (Figure~\ref{fig:headscores}). This highlights heads that potentially hold Task Vectors. We select top heads—$(26, 3)$ and $(11, 3)$—and a lower-ranked head $(27, 5)$. For each, we visualize activation clustering and individual Activation Score per token.

\noindent\textbf{Clustering Visualization.} Highly-ranked heads (e.g., (26,3)) show clear task-based clustering, while low-ranked heads like (27,5) show many small clusters with different tasks, likely based on input semantics. The activations are projected onto 2D using t-SNE~\cite{JMLR:v9:vandermaaten08a} with colors indicating tasks.


\noindent\textbf{Score-per-token Heatmap}. We display ${\rho_{token}}(i)$ values for each token, reflecting spatial positioning in a 2x2 grid. For heads of interest, these values are shown on a heatmap. The CLS token is in the top left, followed by values for quadrants $(x_{s}, y_{s}, x_q, y_q)$. The encoder lacks $y_q$ tokens (bottom right), marked as X. Token values within a head vary widely but show consistency within quadrants, supporting quadrant-based token patching.

\noindent\textbf{Quantitative Clustering Analysis.} To validate our scoring-based clustering quality observations, we report Silhouette and Davies-Bouldin scores for the two highest-ranked heads, two randomly sampled heads, and the two lowest-ranked heads (Table~\ref{tab:metrics_comparison}). Heads scored highly by our method show high-quality clustering scores, while low-scored heads show poor clustering. Randomly selected heads received intermediate scores, further supporting our methodology.

\begin{table}[t]
    \centering
\captionof{table}{\textbf{Task Clustering Quality.} Clustering Scores of Different Attention Heads, ranked by our Activation Score (see Section~\ref{sec:methods}). This indicates that higher Activation Scores indeed correlate with better clustering by tasks.} 
\label{tab:metrics_comparison}
\resizebox{0.5\textheight}{!}{

\begin{tabular}{llcccc}
\toprule
  & (Layer, Head) & Our Score & Silhouette Score & Davies-Bouldin Score \\
& & $\uparrow$ & $\uparrow$ & $\downarrow$ \\
\midrule
High 1 & (26, 3) & \textbf{2.1663} & \textbf{0.3583} & \textbf{1.2744} \\
High 2 & (11, 3) & 1.0827 & 0.2692 & 1.5567 \\
Random 1 & (4, 3) & 0.2329 & 0.0708 & 4.1062 \\
Random 2 & (18, 16) & 0.1259 & 0.0369 & 4.3256 \\
Low 1 & (2, 16) & 0.0221 & -0.0518 & 21.8265 \\
Low 2 & (2, 12) & 0.0264 & -0.0334 & 13.5982 \\
\bottomrule
\end{tabular}
    }
    \end{table}

\subsection{Task Vector Patching}

We present the performance of our task-specific and multi-task methods compared to the original model and baselines. Task Vector interventions improve visual ICL task performance over the original model. Table \ref{tab:results} shows mean and variance across 4 splits (full results in Table \ref{tab:results_full}, Supplementary Materials). Our task-specific models outperform MAE-VQGAN in all tasks, with GRS models surpassing it in some. Our multi-task models excel in all tasks except Segmentation, where they match MAE-VQGAN. GRS multi-task matches the original model, while CMA improves upon random baselines but falls short of our method and GRS.

\begin{table}[t]
    \centering
\captionof{table}{\textbf{Quantitative Analysis.} Results comparison across different tasks and splits, indicating the effectiveness of our task-specific model.}
\label{tab:results}
\small

\resizebox{0.6\textheight}{!}{

\begin{tabular}{lcccc}
\toprule
& Segmentation $\uparrow$ & Lowlight Enhance $\downarrow$ & Colorization $\downarrow$ & In-painting $\downarrow$ \\
Model & (Mean ± STD) & (Mean ± STD) & (Mean ± STD) & (Mean ± STD) \\
\midrule
Original MAE-VQGAN & $0.338 \pm 0.033$ & $0.685 \pm 0.032$ & $0.618 \pm 0.027$ & $0.550 \pm 0.042$\\
\midrule
Random Quadrants & $0.170 \pm 0.061$ & $3.000 \pm 0.967$ & $3.025 \pm 1.190$ & $2.350 \pm 0.955$\\
Random K Layers & $0.090 \pm 0.007$ & $1.825 \pm 0.043$ & $0.568 \pm 0.022$ & $0.875 \pm 0.100$ \\
Top Quadrants & $0.150 \pm 0.023$ & $4.875 \pm 0.228$ & $4.250 \pm 0.269$ & $3.900 \pm 0.141$\\
CMA (Task-specific) & $0.230 \pm 0.012$ & $0.825 \pm 0.043$ & $0.895 \pm 0.063$ & $1.750 \pm 0.112$ \\
CMA (Multi-task) & $0.150 \pm 0.017$ & $1.400 \pm 0.122$ & $1.130 \pm 0.075$ & $1.225 \pm 0.083$ \\
GRS (Task-specific) & $0.320 \pm 0.021$ & $0.600 \pm 0.025$ & $0.555 \pm 0.025$ & $0.580 \pm 0.047$ \\
GRS (Multi-task) & $0.323 \pm 0.019$ & $0.515 \pm 0.032$ & $0.568 \pm 0.029$ & $0.605 \pm 0.036$ \\
\midrule
Ours (Multi-task) & $0.325 \pm 0.026$ & $0.492 \pm 0.025$ & $0.502 \pm 0.036$ & $0.558 \pm 0.022$ \\
Ours (Task-specific) & $\textbf{0.353} \pm 0.028$ & $\textbf{0.458} \pm 0.032$ & $\textbf{0.453} \pm 0.036$ & $\textbf{0.480} \pm 0.022$\\
\bottomrule
\end{tabular}
}   
\end{table}

\begin{figure}
    \centering
     
        \centering
     \includegraphics[height=\textheight]{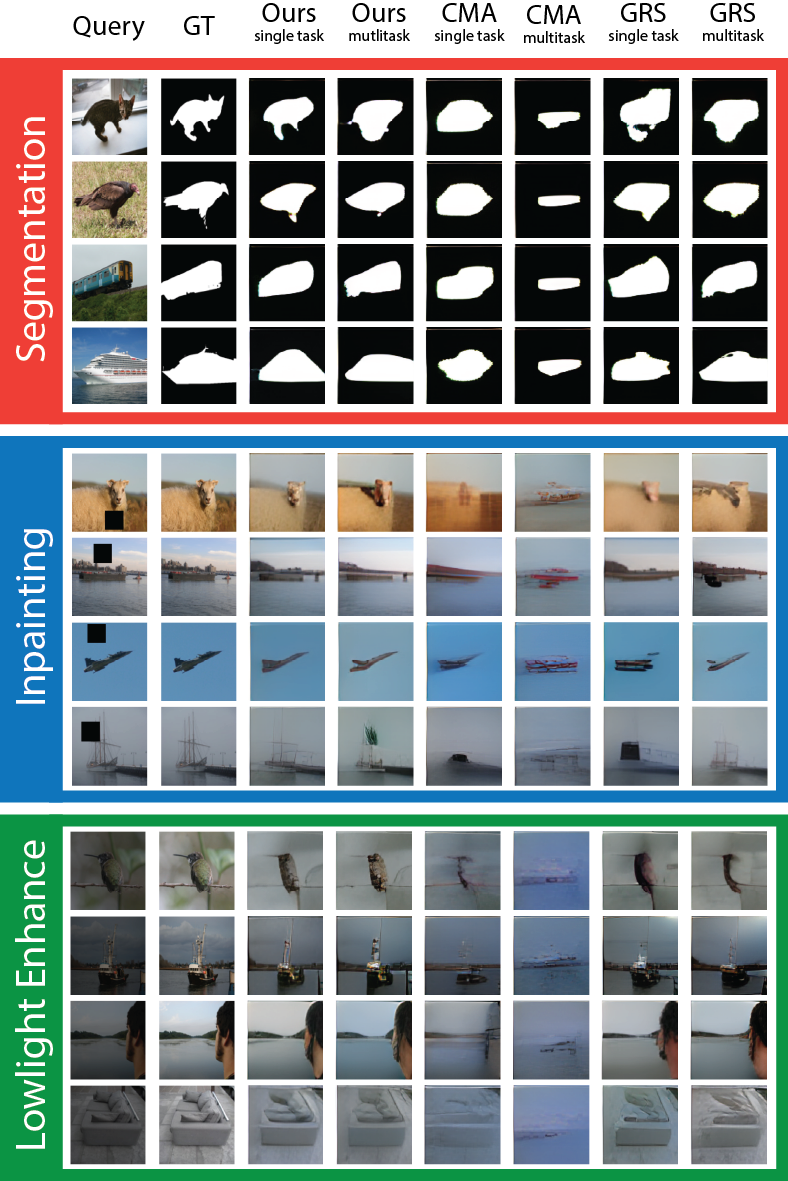}
        \caption{\textbf{Qualitative Examples.} We qualitatively compare the task-specific and multi-task variants of our methodology with the CMA and GRS baselines. Our patching methodology performs better than the original MAE-VQGAN model.}
        \label{fig:example2}
        
\end{figure}

We validate the importance of ranking layers by $\rho_{layer}(l)$ (Section \ref{sec:grs_impl}) and selecting top-k for greedy random search performance. Random K layers generally underperform, except in colorization. Simply patching top quadrants without GRS also yields poor results.
Qualitative results for Segmentation, Lowlight Enhancement, and In-painting are presented. We compare task-specific models of our method with original MAE-VQGAN, CMA, and GRS baselines (Figure~\ref{fig:example}), and visualize task-specific and multi-task variants against CMA and GRS (Figure~\ref{fig:example2}).

\subsection{Ablations}

\noindent\textbf{Task Vectors Location in Encoder vs. Decoder.} We compare interventions isolated to the encoder, decoder, and throughout the whole network. Results show that in-context task learning utilizes both components, with the decoder playing a more crucial role. Intervening in both components is essential for task implementation, supporting our hypothesis of distributed computation with cascading effects throughout the network (see Table~\ref{tab:encoder_decoder}).

\noindent\textbf{Patching Granularity.} Inspired by quadrant patterns in per-token scoring, we explore optimal token grouping to reduce search space dimensionality. Quadrant grouping improves performance for Segmentation and Colorization, while token-level granularity is better for Lowlight Enhancement and In-painting (see Table~\ref{tab:granularity} for mean and variance across 4 splits, and Table~\ref{tab:granularity_full} in Supplementary Materials for individual split evaluations).

\begin{table}[t]
    \centering
\captionof{table}{\textbf{Optimal Patching Granularity.} Patching into Tokens (T), Quadrants (Q), or Heads (H)}
\label{tab:granularity}
\small
\resizebox{\textwidth}{!}{
\begin{tabular}{lcccc}
\toprule
& Segmentation $\uparrow$ & Lowlight Enhance $\downarrow$ & Colorization $\downarrow$ & In-painting $\downarrow$ \\
Model & (Mean ± STD) & (Mean ± STD) & (Mean ± STD) & (Mean ± STD) \\
\midrule
Ours T (Task-specific) & $\textbf{0.350} \pm 0.025$ & $0.495 \pm 0.036$ & $\textbf{0.453} \pm 0.036$ & $0.485 \pm 0.018$\\
Ours Q (Task-specific) & $0.338 \pm 0.023$ & $\textbf{0.458} \pm 0.032$ & $0.465 \pm 0.036$ & $\textbf{0.480} \pm 0.022$ \\
Ours H (Task-specific) & $0.245 \pm 0.015$ & $0.942 \pm 0.070$ & $0.857 \pm 0.069$ & $0.885 \pm 0.067$\\
\midrule
Ours T (Multi-task) & $0.318 \pm 0.023$ & $0.510 \pm 0.028$ & $0.510 \pm 0.039$ & $0.565 \pm 0.025$\\
Ours Q (Multi-task) & $\textbf{0.325} \pm 0.026$ & $\textbf{0.492} \pm 0.025$ & $\textbf{0.502} \pm 0.036$ & $\textbf{0.558} \pm 0.022$\\
Ours H (Multi-task) & $0.253 \pm 0.013$ & $1.105 \pm 0.064$ & $0.998 \pm 0.069$ & $0.980 \pm 0.082$\\
\bottomrule
\end{tabular}
}
\end{table}

\begin{table}[t]

\centering
\captionof{table}{\textbf{Isolating Task Locations.} Patching into Encoder only, Decoder only, and both.}
        \label{tab:encoder_decoder}
\begin{tabular}{lcccc}
\toprule
& \multicolumn{4}{c}{Segmentation $\uparrow$} \\
\cmidrule(lr){2-5}
Model & Split 0 & Split 1 & Split 2 & Split 3  \\
\midrule

Encoder (Task-specific) & 0.09 & 0.14 & 0.14 & 0.13 \\
Decoder (Task-specific) & 0.32 & 0.34 & 0.29 & 0.29 \\
Both (Task-specific) & \textbf{0.35} & \textbf{0.35} & \textbf{0.31} & \textbf{0.29}  \\ 

\bottomrule
\end{tabular}
\end{table}



\section{Limitations}
\label{sec:limit}

While we focused on Task Vectors, other important vector-types might exist, for example, vectors capturing image structure and ordering. We evaluated performance using MSE (except mIoU for Segmentation) after decoding VQGAN tokens. Future work could explore direct evaluation in VQGAN token space using cross entropy loss for potentially more accurate results.

\section{Conclusion}
\label{sec:conc}
In this work we explore the internal mechanisms of visual in-context learning and devise an algorithm to identify Task Vectors, activations present in transformers that can replace the in-context examples to guide the model into performing a specific task. We confirm our approach by adapting MAE-VQGAN to perform tasks without including the ICL demonstration in the prompt by patching the Task Vectors identified. We find that different than in NLP, in computer vision Task Vectors are distributed throughout the network's encoder and decoder.

\clearpage

{\bf Acknowledgements:}
This project has received funding from the European Research Council (ERC) under the European Unions Horizon 2020 research and innovation programme (grant ERC HOLI 819080). Prof. Darrell’s group was supported in part by DoD including DARPA's LwLL and/or SemaFor programs, as well as BAIR's industrial alliance programs. This work was completed in partial fulfillment for the Ph.D degree of the last author.

\bibliographystyle{splncs04}
\bibliography{main}
\clearpage
\section*{Supplementary Material}
\label{sec:supp}
\section{Full Results}

As supplementary material, we provide the tables displaying the full evaluations across the 4 splits of our method alongside the baselines, and the different patching granularities.

\begin{table}[h!]
\centering
\caption{\textbf{Quantitative Analysis.} Results comparison across different tasks and splits, indicating the effectiveness of our task-specific model.}
\label{tab:results_full}
\small
\begin{tabular}{lcccccccc}
\toprule
& \multicolumn{4}{c}{Segmentation $\uparrow$} & \multicolumn{4}{c}{Lowlight Enhancement $\downarrow$} \\
\cmidrule(lr){2-5}
Model & Split 0 & Split 1 & Split 2 & Split 3 & Split 0 & Split 1 & Split 2 & Split 3 \\
\midrule
Original MAE-VQGAN & 0.35 & 0.38 & 0.33 & 0.29 & 0.70 & 0.66 & 0.73 & 0.65 \\
\midrule
Random Quadrants & 0.08 & 0.25 & 0.16 & 0.19 & 4.30 & 2.40 & 3.50 & 1.80 \\
Random K Layers & 0.09 & 0.10 & 0.09 & 0.08 & 1.80 & 1.80 & 1.90 & 1.80\\
Top Quadrants & 0.11 & 0.17 & 0.16 & 0.16 & 4.50 & 5.00 & 5.10 & 4.90 \\
CMA (Task-specific) & 0.23 & 0.25 & 0.22 & 0.22 & 0.76 & 0.83 & 0.88 & 0.83 \\
CMA (Multi-task) & 0.18 & 0.14 & 0.14 & 0.14 & 1.2 & 1.5 & 1.5 & 1.4 \\
GRS (Task-specific) & 0.33 & 0.35 & 0.30 & 0.30 & 0.56 & 0.61 & 0.63 & 0.60 \\
GRS (Multi-task) & 0.33 & 0.35 & 0.31 & 0.30 & 0.47 & 0.52 & 0.56 & 0.51 \\
\midrule
Ours (Multi-task) & 0.35 & 0.35 & 0.31 & 0.29 & 0.46 & 0.49 & 0.53 & 0.49\\
Ours (Task-specific) & \textbf{0.38} & \textbf{0.38} & \textbf{0.33} & \textbf{0.32} & \textbf{0.41} & \textbf{0.46} & \textbf{0.50} & \textbf{0.46} \\

\end{tabular}
\centering
\small
\begin{tabular}{lcccccccc}
\toprule
& \multicolumn{4}{c}{Colorization $\downarrow$} & \multicolumn{4}{c}{In-painting $\downarrow$} \\
\cmidrule(lr){2-5}
Model & Split 0 & Split 1 & Split 2 & Split 3 & Split 0 & Split 1 & Split 2 & Split 3 \\
\midrule
Original MAE-VQGAN & 0.59 & 0.62 & 0.66 & 0.60 & {0.49} & {0.55} & {0.61} & {0.55} \\
\midrule
Random Quadrants & 2.10 & 4.10 & 4.30 & 1.60 & 3.80 & 1.20 & 1.90 & 2.50\\
Random K Layers & 0.54 & 0.57 & 0.60 & 0.56 & 0.72 & 0.89 & 1.00 & 0.89	 \\
Top Quadrants & 3.80 & 4.40 & 4.30 & 4.50 & 3.70 & 3.90 & 4.10 & 3.90 \\
CMA (Task-specific) & 0.79 & 0.92 & 0.96 & 0.91 & 1.6 & 1.8 & 1.9 & 1.7 \\
CMA (Multi-task) & 1.02 & 1.2 & 1.2 & 1.1 & 1.1 & 1.3 & 1.3 & 1.2 \\
GRS (Task-specific) & 0.52 & 0.56 & 0.59 & 0.55 & 0.52 & 0.56 & 0.65 & 0.59\\
GRS (Multi-task) & 0.53 & 0.57 & 0.61 & 0.56 & 0.55 & 0.61 & 0.65 & 0.61 \\
\midrule
Ours (Multi-task) & 0.45 & 0.51 & 0.55 & 0.50 & 0.53 & 0.56 & 0.59 & 0.55\\
Ours (Task-specific) & \textbf{0.40} & \textbf{0.46} & \textbf{0.50} & \textbf{0.45} & \textbf{0.45} & \textbf{0.49} & \textbf{0.51} & \textbf{0.47} \\ 

\bottomrule
\end{tabular}
\end{table}

\begin{table}[h!]
\centering
\caption{\textbf{Optimal Patching Granularity.} Patching into Tokens (T), Quadrants (Q), or Heads (H)}
\label{tab:granularity_full}
\small
\begin{tabular}{lcccccccc}
\toprule
& \multicolumn{4}{c}{Segmentation $\uparrow$} & \multicolumn{4}{c}{Lowlight Enhancement $\downarrow$} \\
\cmidrule(lr){2-5}
Model & Split 0 & Split 1 & Split 2 & Split 3 & Split 0 & Split 1 & Split 2 & Split 3 \\
\midrule
T (Task-specific) & \textbf{0.38} & \textbf{0.37} & \textbf{0.33} & \textbf{0.32} & 0.44 & 0.50 & 0.54 & 0.50 \\
Q (Task-specific) & 0.36 & 0.36 & 0.32 & 0.31 & \textbf{0.41} & \textbf{0.46} & \textbf{0.50} & \textbf{0.46} \\
H (Task-specific) & 0.24 & 0.27 & 0.23 & 0.24 & 0.83 & 0.97 & 1.02 & 0.95 \\
\midrule
T (Multi-task) & 0.34 & 0.34 & 0.30 & 0.29 & 0.47 & 0.51 & 0.55 & 0.51\\
Q (Multi-task) & \textbf{0.35} & \textbf{0.35} & \textbf{0.31} & \textbf{0.29} & \textbf{0.46} & \textbf{0.49} & \textbf{0.53} & \textbf{0.49}\\
H (Multi-task) & 0.26 & 0.27 & 0.24 & 0.24 & 1.01 & 1.12 & 1.19 & 1.10\\

\bottomrule
\end{tabular}
\centering
\small
\begin{tabular}{lcccccccc}
\toprule
& \multicolumn{4}{c}{Colorization $\downarrow$} & \multicolumn{4}{c}{In-painting $\downarrow$} \\
\cmidrule(lr){2-5}
Model & Split 0 & Split 1 & Split 2 & Split 3 & Split 0 & Split 1 & Split 2 & Split 3 \\
T (Task-specific) & \textbf{0.40} & \textbf{0.46} & \textbf{0.50} & \textbf{0.45} & 0.46 & 0.49 & 0.51 & 0.48 \\ 
Q (Task-specific) & 0.41 & 0.47 & 0.51& 0.47 & \textbf{0.45} & \textbf{0.49} & \textbf{0.51} & \textbf{0.47} \\ 
H (Task-specific) & 0.75 & 0.88 & 0.94 & 0.86 & 0.78 & 0.92 & 0.96 & 0.88 \\
\midrule
T (Multi-task) & 0.45 & 0.51 & 0.56 & 0.52 & 0.53 & 0.57 & 0.60 & 0.56\\
Q (Multi-task) & \textbf{0.45} & \textbf{0.51} & \textbf{0.55} & \textbf{0.50} & \textbf{0.53} & \textbf{0.56} & \textbf{0.59} & \textbf{0.55}\\
H (Multi-task) & 0.89 & 1.02 & 1.08 & 1.0 & 0.85 & 1.02 & 1.07 & 0.98\\

\bottomrule
\end{tabular}
\end{table}

\section{Additional Explorations}

We include a handful of additional experiments that serve as potential future avenues for research.

\noindent\textbf{Finding Task Vectors in Other Architectures}.
We evaluate our method to find Task Vectors on Llama2 7B, an autoregressive decoder-only architecture on NLP tasks and compare the Top-1 accuracy to previous reported results by Todd et al.~\cite{todd2023function}. Our method performs better than previous approaches, beating 10-shot in 2 tasks while reducing FLOPs by 92.5\% (Table \ref{tab:llama2_comparison}).

\begin{table}[h]
\centering
\caption{Comparison of Task Vector methods on Llama2 7B for various NLP tasks}
\label{tab:llama2_comparison}
\begin{tabular}{l|c|c|c|c}
Method  & Landmark to Country & Present to Past & Country to Capital\\ \hline
0-shot & 0.0 & 0.084 & 0.047\\
2-shot & 0.868 & 0.95 & 0.92 \\
10-shot & \textbf{0.88} & 0.967 & 0.951 \\
0-shot + CMA &  0.691 & 0.88 & 0.825 \\
0-shot+ Ours & 0.8629 & \textbf{0.983} & \textbf{0.9524} \\ 
\end{tabular}
\end{table}

\begin{figure}[h]
\centering
\includegraphics[width=0.6\columnwidth]{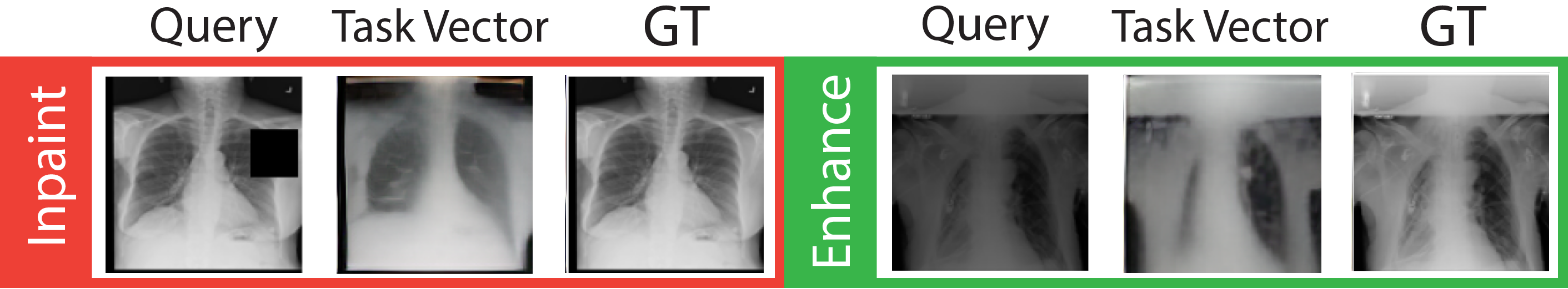}
\caption{Out of domain evaluation on x-ray images.}
\label{fig:xray_results}
\end{figure}

\noindent\textbf{Out of Domain Evaluation}. We evaluate the performance of in-painting and low-light enhancement task-specific models on x-ray images. The qualitative results (Figure \ref{fig:xray_results}) show proper task implementation. The resulting x-ray images are slightly more blurry, which is due to a limitation in the underlying VQGAN tokenizer, previously reported in Bar et al.~\cite{bar2022visual}.

\begin{figure}[h]
\centering
\includegraphics[width=0.5\columnwidth]{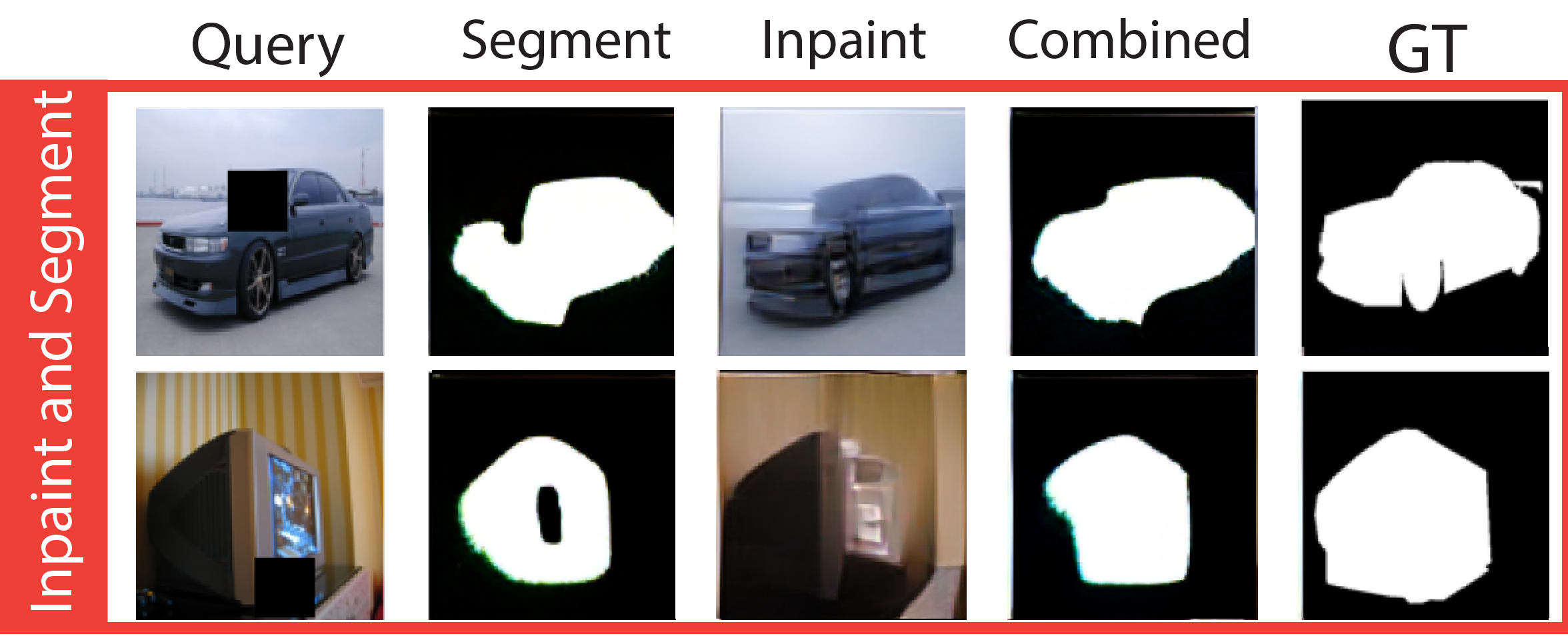}
\caption{Task vector arithmetic results.}
\label{fig:task_arithmetic}
\end{figure}

\noindent\textbf{Task Vector Arithmetic}. We explore whether vector arithmetic can be performed on the task representations to create new tasks by composition. By utilizing the multitask patching positions, we can compose the tasks of in-painting and segmentation by combining their individual Task Vectors as follows: $combined = inpaint + segment - identity$ similarly to Todd et al.~\cite{todd2023function}, where identity is the identity mapping task described in the paper. We compute these new mean activations and patch them into the previously determined multitask positions, and evaluate using as input a masked image and its corresponding segmentation. Figure \ref{fig:task_arithmetic} shows that the tasks are indeed composable, where $combined$ performs qualitatively better than $segment$ which suffers from holes.

\noindent\textbf{One-shot Task Vectors}. We investigate whether 1-shot performance can be improved with Task Vectors. Table \ref{tab:one_shot_tv} shows that additional in-context examples (or shots) are not necessary when using Task Vectors. Intuitively, Task Vectors already lead to improved performance and additional in-context learning examples do not add more information.

\begin{table}[h]
\centering
\caption{Comparison of no ICL demonstration and one-shot performance with Task Vectors (TV)}
\label{tab:one_shot_tv}
\begin{tabular}{l|c|c|c|c}
Method & Segm & LowLight & Inpaint & Colorize\\
 &mIoU&MSE&MSE&MSE \\ \hline
Original one-shot & 0.338 & 0.6 & 0.55  & 0.618 \\
No ICL + TV & \textbf{0.353} & \textbf{0.458} & 0.480  & \textbf{0.453}  \\
One-shot + TV & 0.346 & 0.496 & \textbf{0.454} & 0.490 \\
\end{tabular}
\end{table}

\section{More Qualitative Examples}

We provide a wider selection of examples comparing our task vector patching methodology in comparison to \textbf{1) the original one-shot MAE-VQGAN, CMA, and GRS, and 2) our ablations.} Alongside each figure, we accompany it with an according analysis.

\begin{figure*}[th!]
    \centering
    \includegraphics[height=0.83\textheight]{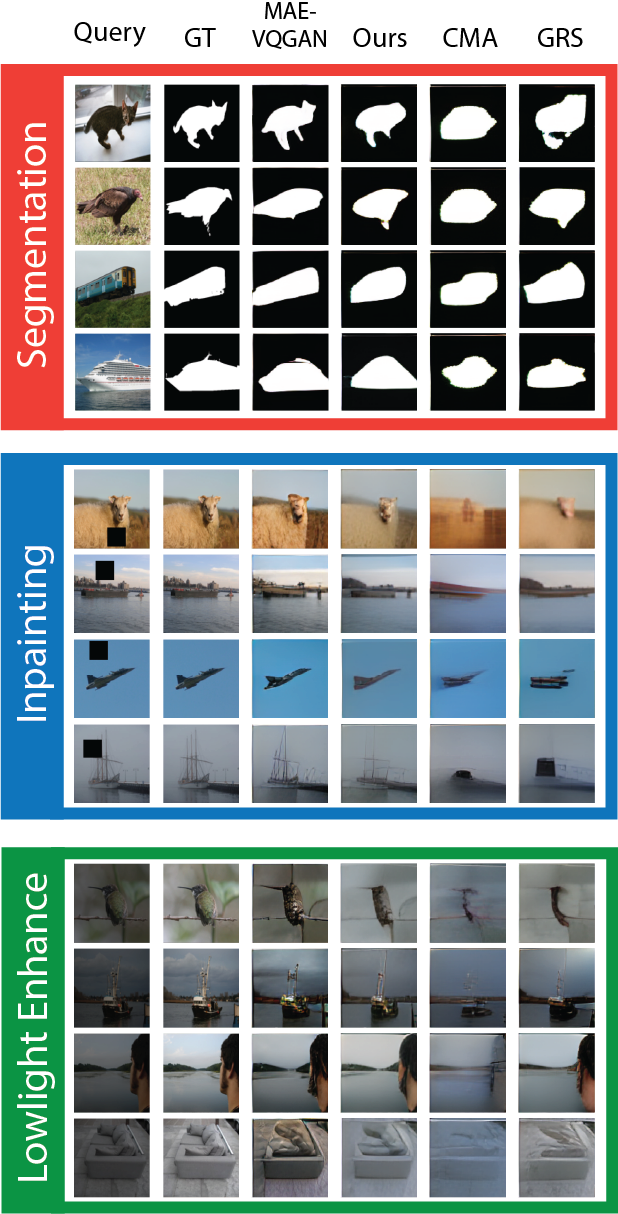}
    \caption{\textbf{Qualitative Examples.} We qualitatively compare our the task-specific variants of our methodology with the original model and the CMA and GRS baselines.}
    \label{fig:supp_example}
\end{figure*}
\begin{figure*}[th!]
    \centering
    \includegraphics[height=0.85\textheight]{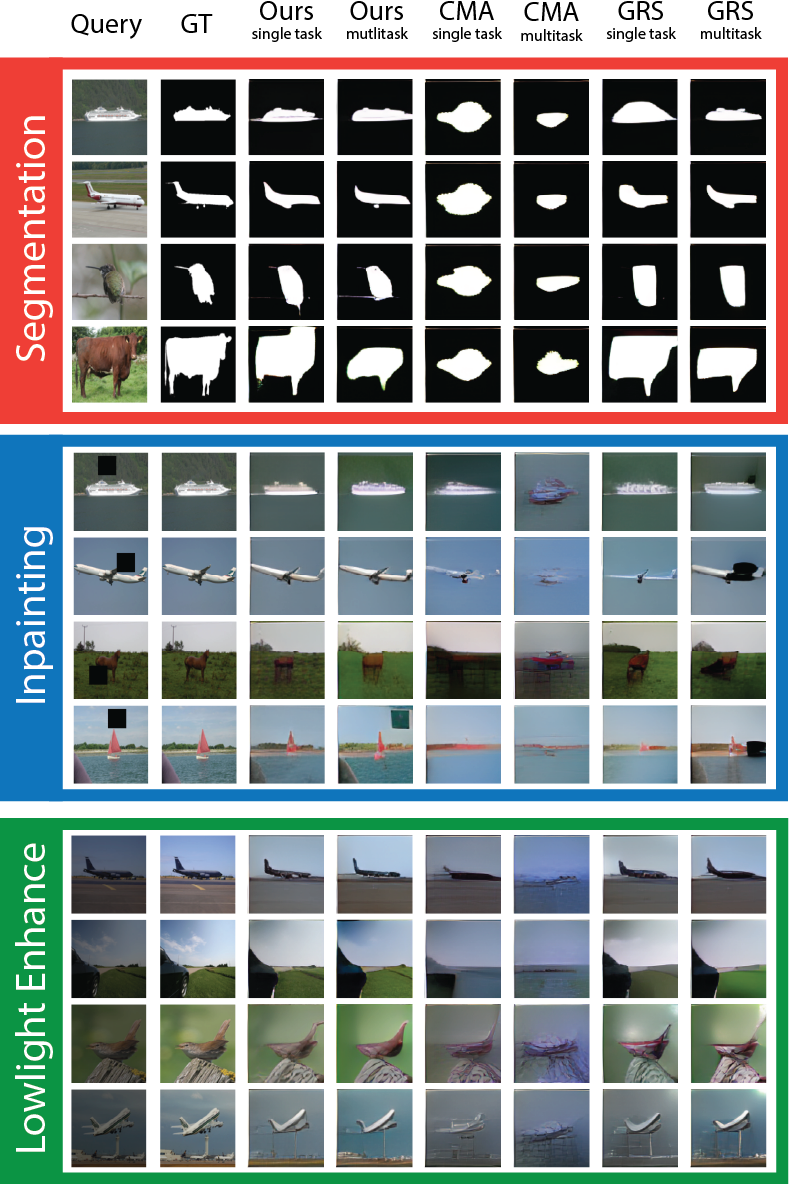}
    \caption{\textbf{Qualitative Examples.} We qualitatively compare the task-specific and multi-task variants of our methodology with the CMA and GRS baselines.}
    \label{fig:supp_example2}
\end{figure*}

 First, we visualize the task-specific models of our methodology alongside the original MAE-VQGAN, and the CMA and GRS baselines (see Figure~\ref{fig:supp_example}). Secondly, we visualize the task-specific and multi-task variants of our methodology in comparison to CMA (see Figure~\ref{fig:supp_example2}).

\begin{figure*}[th]
    \centering
    \includegraphics[width=\linewidth]{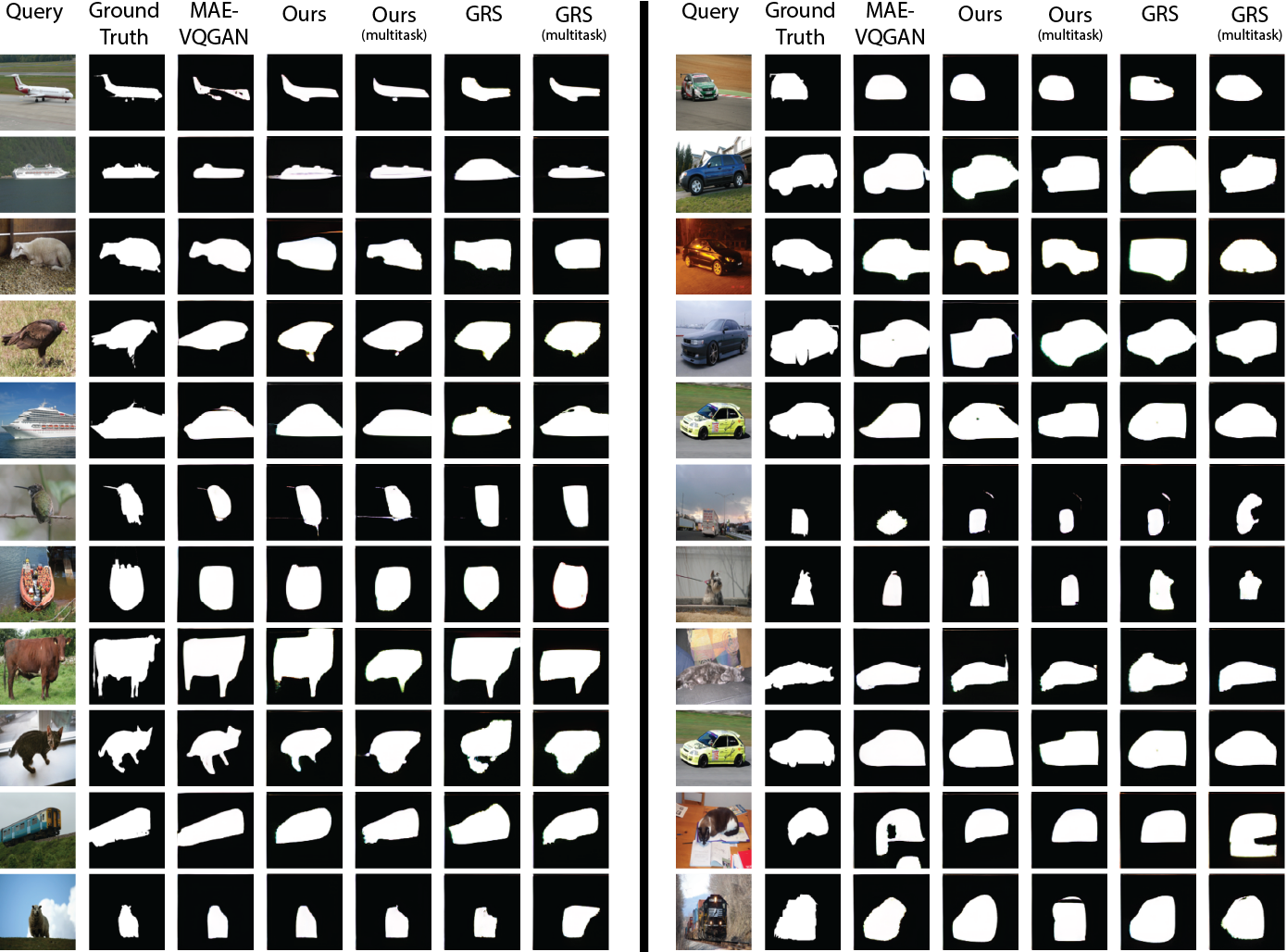}
    \caption{\textbf{Our Results on Segmentation Task}}
    \label{fig:seg_suppl_pg}
\end{figure*}

\textbf{Qualitative Analysis for Segmentation.} In Figure \ref{fig:seg_suppl_pg}, we compare our methodology and GRS task-specific and multi-task methods to the original one-shot MAE-VQGAN performance on the task of Segmentation. It appears that our method is good at segmenting the coarse and fine details of the object of focus. In many cases, the segmentations generated by the original MAE-VQGAN suffer from holes or incomplete masks. In contrast, our method outputs consistent and coherent masks. On the other hand, the GRS method suffers with particular details especially observable when attempting to segment an animal's ear or leg. However, in many such cases it performs better than MAE-VQGAN at getting the general shape of objects.

\begin{figure*}[!ht]
    \centering
    \includegraphics[height=0.45\textheight]{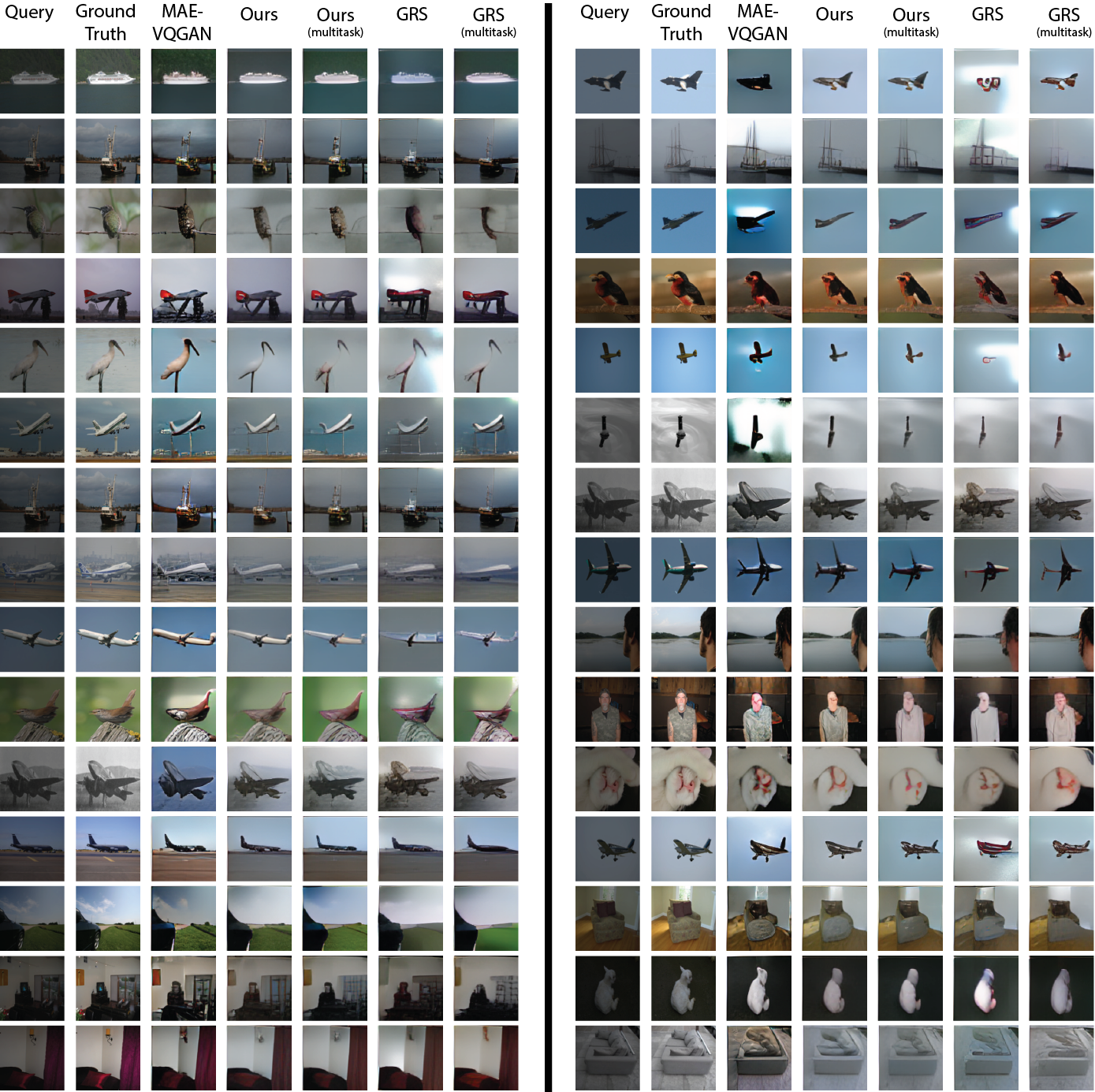}
    \caption{\textbf{Our Results on Lowlight Enhancement Task}}
    \label{fig:low_suppl_pg}
\end{figure*}

\textbf{Qualitative Analysis for Lowlight Enhancement.} In Figure \ref{fig:low_suppl_pg}, we compare our methodology and GRS task-specific and multi-task methods to the original one-shot MAE-VQGAN performance on the task of Lowlight Enhancement. It appears that the GRS method suffers in maintaining the visual qualities of the query image. However there are many cases where MAE-VQGAN assigns bright colors which is likely due to the particular prompt in use and the inherent ambiguities of the task. On the other hand, our method--particularly the multi-task variant--ouputs consistently better results with accurate visual qualities. In some cases our method produces somewhat muted or blurry results which may be a consequence of using MSE in the pixel space as supervision, but nonetheless reports better quantitative performance.

\begin{figure*}[!ht]
    \centering
    \includegraphics[height=0.43\textheight]{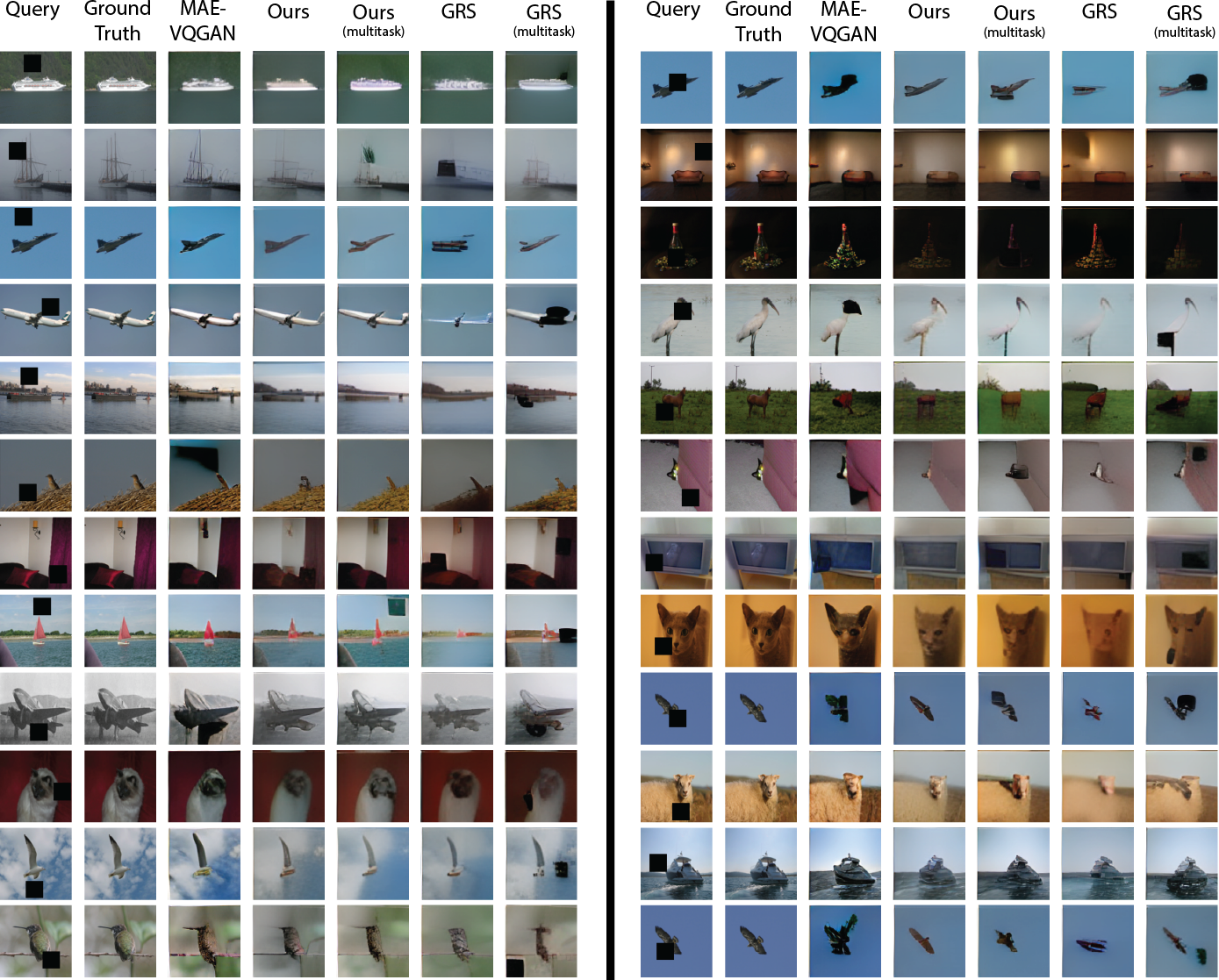}
    \caption{\textbf{Our Results on In-painting Task}}
    \label{fig:inpaint_suppl_pg}
\end{figure*}

\textbf{Qualitative Analysis for In-painting.} In Figure \ref{fig:inpaint_suppl_pg}, we compare our methodology and GRS task-specific and multi-task methods to the original one-shot MAE-VQGAN performance on the task of In-painting. We observe that our method consistently outperforms the original model. However, it appears that the GRS task-vector patching method--once again--suffers in maintaining the higher frequency components of the query image; it appears to reduce the contrast of the image and reduce saturation. However, there are many such cases where the original MAE-VQGAN one-shot technique fails to appropriately implement the task while our method succeeds. The original model's performance depends heavily on the specific prompt used which may be the root cause of failures while task-vector patching succeeds.

\begin{figure*}[!ht]
    \centering
    \includegraphics[height=0.83\textheight]{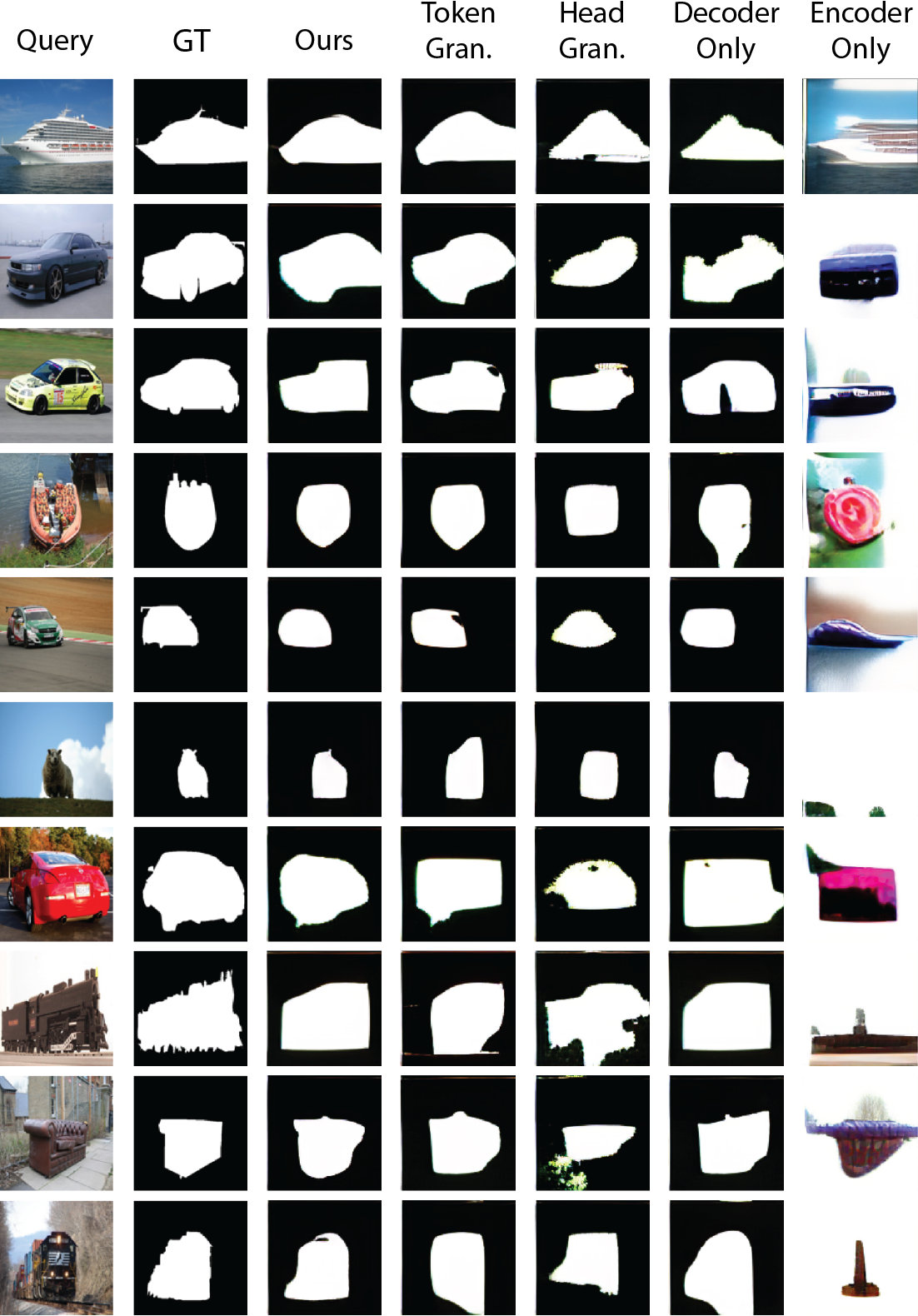}
    \caption{\textbf{REINFORCE Ablations for Segmentation Task}}
    \label{fig:supp_ablation_pg}
\end{figure*}

\textbf{Qualitative Analysis of Ablations.} Finally, we present the qualitative analysis of the different ablations for the Segmentation task in Figure \ref{fig:supp_ablation_pg}. The benefits of observing the visual features of the different ablations in addition to quantitative analysis becomes clear when comparing the Decoder Only and Encoder Only columns. Here it is clear that patching into decoder is of utmost importance in relation to patching into the encoder; the distinction is clear when observing qualitative features. In the end, it is both parts in synchrony which allow for the implementation of in-context learning.

\section{Greedy Random Search baseline}
\subsection{Selecting Task Vectors via Greedy Random Search} 
\label{sec:grs_impl}
For every task $j$ we apply the following algorithm to obtain a task-specific model.

\vspace{1.3mm}

\noindent\textbf{Input.} The mean activations $\{\mu_{i,j}\}$, pretrained visual prompting model $F(\cdot)$, an evaluation set, and the aggregated per-layer score defined as $\rho_{layer}(l)$:

\begin{align}
\rho_{layer}(l) &= \sum_{m,k} \rho_{token}(i=(l,m,k))
\end{align}

\noindent\textbf{Initialization.} 
Initialize a set of binary indicators $\{\alpha_{i,j}\}$ for all $i$, where $\alpha_{i,j} \in \{0,1\}$ signifies whether the mean task activation $\mu_{i,j}$ is a task vector. Next we describe the algorithm to choose the values of $\alpha_{i,j}$.

For every activation $i=(l,m,k)$ in the top $k$ scoring layers w.r.t $\rho_{layer}(l)$, randomly choose the value of $\alpha_i$ by sampling from a Bernoulli random variable with parameter value $p$. Set the activation vectors to be $z_j=\{\mu_{i,j} |\forall ij ~\text{if}~\alpha_{i,j}=1\}$. Evaluate the now task-specific model $F(\cdot|z_j)$ on a held-out validation set.
Run for $T$ trials and for every $i,j$ set the values of $\alpha_{i,j}$ to be the values from the most successful trial.

\noindent\textbf{Greedy Search.} Iterate over $l$ in the top $k$ scoring layers sorted by $\rho_{layer}(l)$ from high to low, pick activation $i=(l,m,k)$, flip the value $\alpha_{i,j}$ and evaluate the validation score for $F(\cdot|z_j)$. After having evaluated each flip of the $\alpha_{i,j}$ on the particular layer $l$ we keep the $\alpha_{i,j}$ which performed best (based on a certain evaluation function) or continue to the next layer if the performance did not improve.

\noindent\textbf{Termination.} When one search loop across the $k$ layers results in no changes - or after $10k$ iterations, the search has thus converged and we return the single-task model $F(\cdot|z_j)$.

This procedure is outlined for every token, attention head, and layer. However, it is possible to apply it in different levels of granularity. For example, by patching group of tokens from the same quadrant, patching all the tokens in an attention head, or patching the entire layer. We discuss these design choices in the next section.

\subsection{Greedy Random Search Implementation Experiments}
In this section we describe the set of experiments conducted to ascertain the particular implementation details of the greedy random search, validating the design choices.

\vspace{1.3mm}
\noindent\textbf{Implementation Details} We search through the top $k=17$ layers ranked by Activation Scoring. During the initialization phase we sample $\alpha_{i,j} \in \{0,1\}$ from a Bernoulli distribution with a parameter of $p=0.3$ (probability of selecting 1) and evaluate performance. We repeat this for $T=100$ trials and continue with the best performing $\alpha_{i,j}$. 
Furthermore, we perform a grouping of token positions $i$ in each individual attention head into 3 groups: the CLS token, bottom left quadrant, and bottom right quadrant. This serves to further reduce the search space. We use a set of 10 training images to supervise the search. These design decisions are further validated through ablation experiments. 


\vspace{1.3mm}
\noindent\textbf{Selecting Initialization Parameters.} 
For the initialization of the Greedy Random Search there are two parameters, $k$ which dictates how many layers to search across and the Bernoulli random variable parameter $p$ which dictates the probability at which we set $\alpha_{i,j}$ to be 1 during the initialization phase. The question is, which $k$ value is best at narrowing down the search space without restricting our ability to induce task implementation, and what is the best according $p$ value? We ascertain this by searching for the optimal configuration to initialize the Greedy Random Search. We perform a grid search for $k$ values from 14 to 20, and $p$ values from 0.1 to 0.6 and report the evaluation metric for the Segmentation task on the batch of 10 images. Our goal is to find the $(k,p)$ pair with highest performing random initialization.

\vspace{1.3mm}
\noindent\textbf{Task Vectors Location in Encoder vs. Decoder.} 
Similar to the ablation conducted for our main methodology (via REINFORCE~\cite{williams1992simple}) implementation, we execute the Greedy Random Search for the Segmentation task by restricting interventions to the encoder only, the decoder only, and allowing for interventions throughout the whole network. It is key to note that in order to restrict interventions to the decoder only, which has 8 layers, the $k$ value must be set to 8, whereas for isolating the encoder we can keep the original $k=17$ value. We report the mIoU on the four splits for the Segmentation task seeking to find if interventions in both parts of the model are required for appropriate task implementation. 

\vspace{1.3mm}
\noindent\textbf{Patching Granularity.} We execute our Greedy Random Search with three granularity levels, grouping by Quadrants, grouping by Heads, and grouping by Layers, and report the mIoU performance on the four splits for the Segmentation task.

\subsection{GRS Implementation Experiments Results}
 
\begin{figure*}
    \centering
    \includegraphics[width=0.8\linewidth]{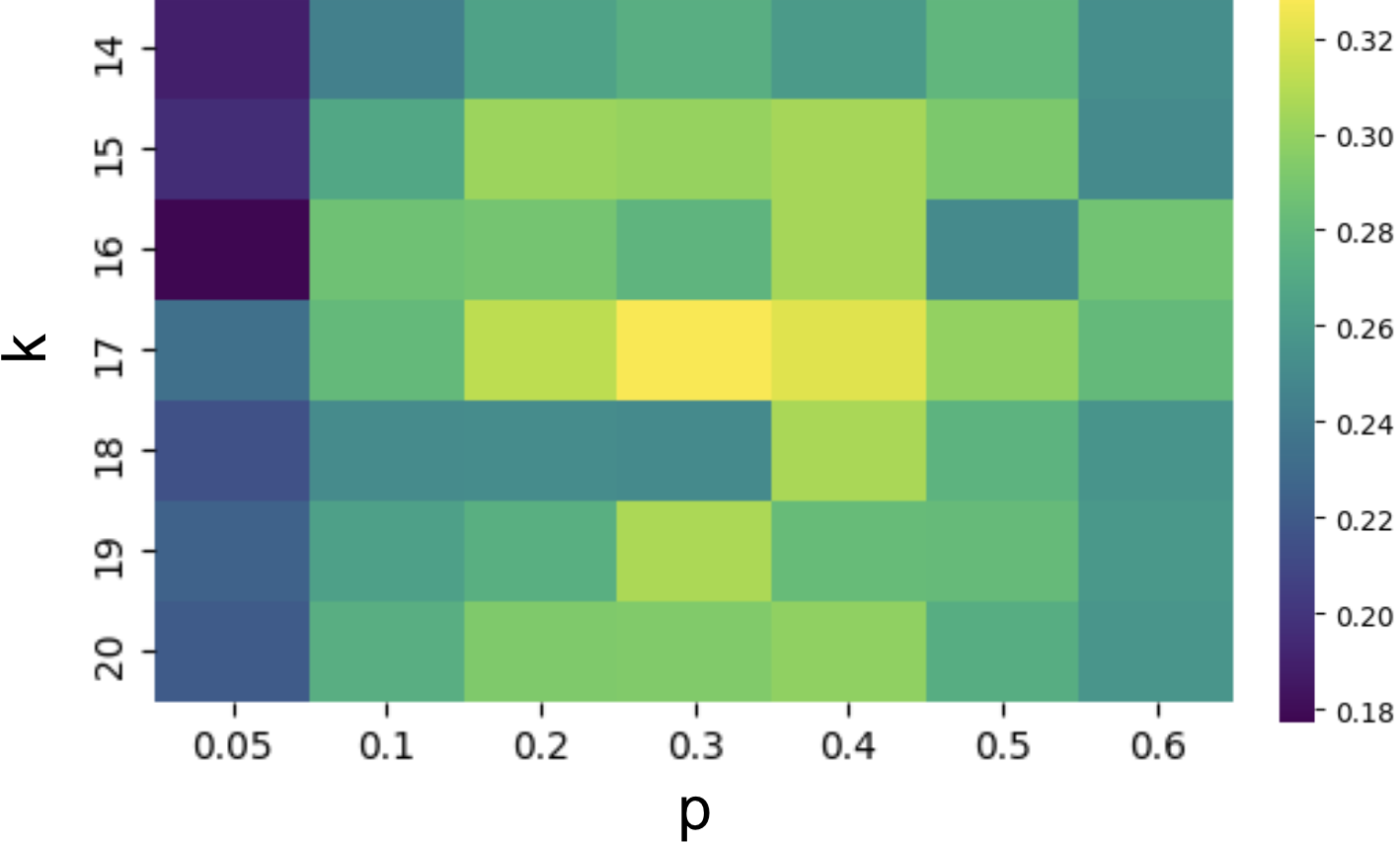}
    \caption{\textbf{Selecting Initialization Parameters.} We evaluate Foreground Segmentation mIoU on Pascal 5i using $10$ images for different random initialization parameterized by $K$ and $p$.}
    \label{fig:initi_ablation_grs}
\end{figure*}

\vspace{1.3mm}
\noindent\textbf{Selecting K.} We explore the optimal parameters for a random initialization. We find the best setup to be to constrain the search across the top $k=17$ layers, sampling quadrants to patch with a probability of $p=0.3$ (see Figure~\ref{fig:initi_ablation_grs}).

\vspace{1.3mm}
\noindent\textbf{Task Vectors Location in Encoder vs. Decoder.} We report the results on isolating the set of possible interventions to the encoder only, decoder only, in contrast to allowing interventions throughout the whole network. We can observe that in-context task learning builds upon both model components. The decoder, however, is more important. It is clear that intervening in both components is crucial for task implementation as we hypothesize that it is computed in a distributed fashion with cascading higher-order effects through the network where early interventions have strong downstream effects (see Table~\ref{tab:encoder_decoder_grs}). 

\begin{table}[htbp]
    \begin{minipage}[t]{0.47\textwidth}
        \centering
\caption{\textbf{Isolating Task Locations.} Patching into Encoder only, Decoder only, and Both}
        \label{tab:encoder_decoder_grs}
\resizebox{\textwidth}{!}{%
\begin{tabular}{lcccc}
\toprule
& \multicolumn{4}{c}{Segmentation $\uparrow$} \\
\cmidrule(lr){2-5}
Model & Split 0 & Split 1 & Split 2 & Split 3  \\
\midrule

GRS Both (Task-specific) & \textbf{0.33} & \textbf{0.35} & \textbf{0.30} & \textbf{0.30}  \\
GRS Encoder (Task-specific) & 0.13 & 0.22 & 0.20 & 0.20 \\
GRS Decoder (Task-specific) & 0.26 & 0.28 & 0.25 & 0.25 \\
\bottomrule
\end{tabular}}
    \end{minipage}
    \hfill
    \begin{minipage}[t]{0.48\textwidth}
        \centering
\caption{\textbf{Optimal Patching Granularity.} Patching into Full Layers, Full Heads, or Quadrants only}
        \label{tab:granularity_grs}
\resizebox{\textwidth}{!}{%
\begin{tabular}{lcccc}
\toprule
& \multicolumn{4}{c}{Segmentation $\uparrow$} \\
\cmidrule(lr){2-5}
Model & Split 0 & Split 1 & Split 2 & Split 3  \\
\midrule
GRS Quadrants (Task-specific) & \textbf{0.33} & \textbf{0.35} & \textbf{0.30} & \textbf{0.30}  \\
GRS Heads (Task-specific) & 0.15 & 0.15 & 0.14 & 0.13\\
GRS Layers (Task-specific) & 0.28 & 0.31 & 0.26 & 0.27 \\
\bottomrule
\end{tabular}}
    \end{minipage}
\end{table}



\vspace{1.3mm}
\noindent\textbf{Patching Granularity.} We explore the optimal granularity at which to group the tokens to reduce the dimensionality of the search space. Motivated by the emergence of quadrants in the per-token scoring visualization, and validated by attempting to group by whole attention heads (patching into all the tokens in the attention head) and group by whole layers (patching into all the attention heads of a layer), it is clear that quadrants provide the best trade-off between reducing the dimensionality of the search space and performance  (see Table~\ref{tab:granularity_grs}). It is interesting to note that patching into full layers reduces the search space to a size of $2^{\text{32}}$ whereas attention heads is $2^{\text{512}}$ and quadrants is $2^{\text{768}}$ for the encoder and $2^{\text{384}}$ in the decoder (disregarding top-k layer selection). 

\subsection{GRS Baseline Qualitative Comparisons}

We provide a wider selection of examples comparing our GRS task vector patching methodology in comparison to \textbf{1) a selection of baselines, and 2) our ablations.} Alongside each figure, we accompany it with an according analysis.

\begin{figure*}[!ht]
    \centering
    \includegraphics[width=0.95\linewidth]{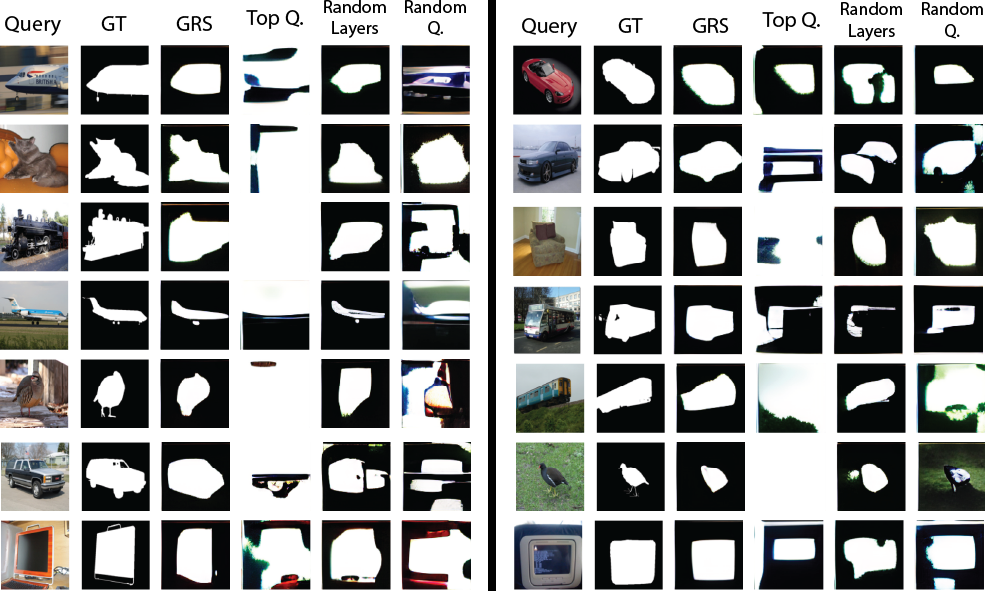}
    \caption{\textbf{GRS Baseline Comparison for Segmentation}}
    \label{fig:segm_baseline}
\end{figure*}

\begin{figure*}[!ht]
    \centering
    \includegraphics[width=0.95\linewidth]{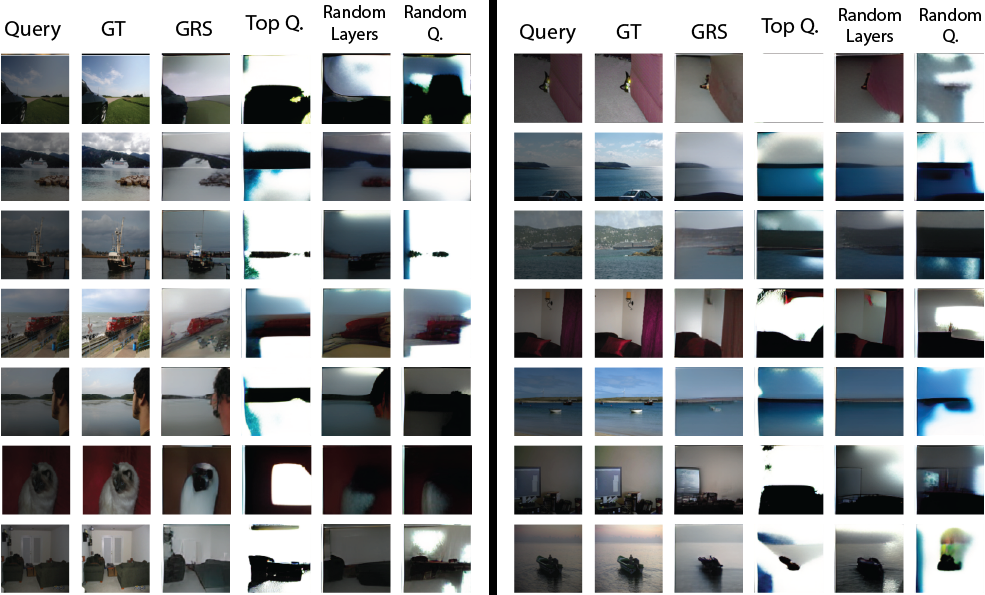}
    \caption{\textbf{GRS Baseline Comparison for Low light Enhancement}}
    \label{fig:lowlight_baseline}
\end{figure*}

\begin{figure*}[!ht]
    \centering
    \includegraphics[width=\linewidth]{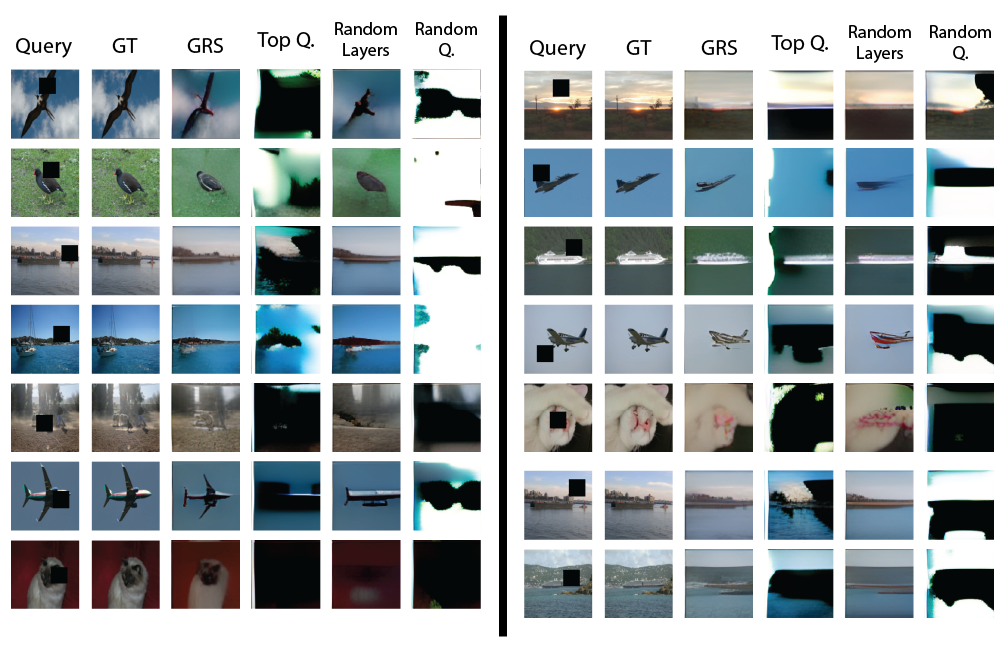}
    \caption{\textbf{GRS Baseline Comparison for In-painting}}
    \label{fig:inpaint_baseline}
\end{figure*}

In the Figures \ref{fig:segm_baseline}, \ref{fig:lowlight_baseline}, and \ref{fig:inpaint_baseline} we compare our GRS task-specific method with a handful of baselines defined in section \ref{sec:vector_patching_exp}. We have abbreviated Top Quadrants as Top Q, Random K Layers as Random Layers, and Random Quadrants as Random Q. It is clear that Top Quadrants struggles to output coherent completions. We believe this to be because of the need of patching into positions of different purposes other than task implementation such as positions that encode the input-output structure that a one-shot prompt provides. Further opportunities for exploration could include other scoring terms that take into account structural information provided by different prompt orientations. Random K Layers performs surprisingly well due to the efficiency of the Greedy Random Search but nonetheless does not reach the performance of using our scoring mechanism to select the top $K$ layers. Finally Random Quadrants struggles to complete coherent results. 

\begin{figure*}[!ht]
    \centering
    \includegraphics[height=\textheight]{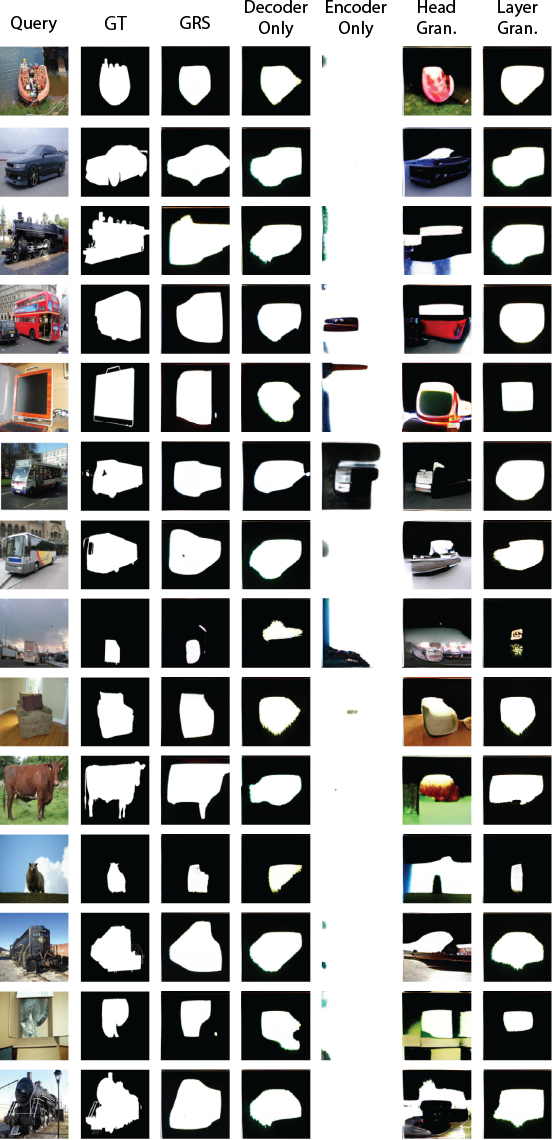}
    \caption{\textbf{GRS Ablation for Segmentation Task}}
    \label{fig:supp_ablation}
\end{figure*}

Finally, we present the qualitative analysis of the different ablations for the Segmentation task in Figure \ref{fig:supp_ablation}. Similarly to our main method (via REINFORCE~\cite{williams1992simple}), it is clear that patching into decoder is more important than patching into the encoder but in the end, it is both parts in synchrony which report the best performance. Furthermore, it is clear that the optimal granularity for patching is at a quadrant level. We find it counterintuitive that layer-level patching performs better than head-level patching--as one would assume that a finer granularity provides better accuracies. However, we believe that by grouping per-layer we significantly reduce the search space (by a factor of 16) which reduces the probability of falling into a local optimum; whereas grouping by head, we suffer from a reduced precision but do not gain the benefits of a reduced  search space magnitude (factor of 2 for encoder and factor of 3 for decoder when grouping by head instead of 16 when grouping by layer). Further exploration in this direction is of interest. 
\end{document}